\documentclass{article}

\usepackage{mathrsfs}
\usepackage{amsmath}
\usepackage{amsfonts}
\usepackage{amssymb}
\usepackage{commath}
\usepackage{graphicx,color}
\usepackage{url}
\usepackage{latexsym,enumerate}
\usepackage{booktabs}
\usepackage{framed}
\usepackage{subfigure}
\usepackage{multirow} 
\usepackage{color}
\usepackage{colortbl}

\usepackage{algorithm}
\usepackage{algorithmic}

\definecolor{gray1}{rgb}{0.8,0.8,0.8}
\definecolor{gray2}{rgb}{0.95,0.95,0.95}

\newcommand{\RE}{\,{\rm Re}}

\newcommand{\argmin}{\mathop{\rm argmin}}

\newcommand{\Shrink}{\mathop{\rm Shrink}}

\newcommand{\CST}{\mathop{\rm CST}}

\newcommand{\cC}{{\mathcal C}}

\newcommand{\cF}{{\mathcal F}}

\newcommand{\cT}{{\mathcal T}}

\newcommand{\Bf}{{\mathbf f}}
\newcommand{\Bu}{{\mathbf u}}

\newcommand{\Br}{{\mathbf r}}
\newcommand{\Bv}{{\mathbf v}}
\newcommand{\Be}{{\mathbf e}}

\newcommand{\Bw}{{\mathbf w}}

\newcommand{\Bg}{{\mathbf g}}

\newcommand{\Beps}{{\boldsymbol \epsilon}}

\newcommand{\Bome}{{\boldsymbol \omega}}

\newcommand{\Bk}{{\boldsymbol k}}
\newcommand{\Bz}{{\boldsymbol z}}

\newcommand{\BDm}{{\mathbf{D_m}}}

\newcommand{\BDnT}{{\mathbf{D}_{\mathbf n}^{\text T}}}

\begin{document}

\title{Simultaneous Inpainting and Denoising by Directional Global Three-part Decomposition: Connecting Variational and Fourier Domain Based Image Processing}

\author{D.H. Thai\thanks{Institute for Mathematical Stochastics, 
University of Goettingen,
Goldschmidtstr. 7, 37077 G\"ottingen, Germany,
and Statistical and Applied Mathematical Science Institute (SAMSI), USA.
Email: dhthai@samsi.info}
\ and C. Gottschlich\thanks{Institute for Mathematical Stochastics, 
University of Goettingen,
Goldschmidtstr. 7, 37077 G\"ottingen, Germany.
Email: gottschlich@math.uni-goettingen.de}}

\date{}

\maketitle

\begin{abstract}

We consider the very challenging task 
of restoring images  
(i)~which have a large number of missing pixels, 
(ii)~whose existing pixels are corrupted by noise and 
(iii)~the ideal image to be restored contains both cartoon and texture elements.
The combination of these three properties makes this inverse problem 
a very difficult one.
The solution proposed in this manuscript is based on 
directional global three-part decomposition (DG3PD) \cite{ThaiGottschlich2015DG3PD}
with directional total variation norm, directional G-norm and $\ell_\infty$-norm in curvelet domain
as key ingredients of the model.
Image decomposition by DG3PD 
enables a decoupled inpainting and denoising 
of the cartoon and texture components.
A comparison to existing approaches for inpainting and denoising 
shows the advantages of the proposed method.
Moreover, 
we regard the image restoration problem from the viewpoint 
of a Bayesian framework and 
we discuss the connections between the proposed 
solution by function space and 
related image representation by harmonic analysis and pyramid decomposition.

\end{abstract}

\section*{Keywords}

Image decomposition, variational calculus, inverse problems,
image inpainting, image denoising, 
cartoon image, texture image, noise, residual image,
feature extraction.

\section{Introduction and Related Work} \label{sec:Introduction}

Image enhancement and image restoration are two superordinate concepts 
in image processing which encompass a plethora of methods 
to solve a multitude of important real-world problems \cite{DigitalImageProcessing2002,Szeliski2011}.
Image enhancement has the goal of improving an input image for a specific application,
e.g. in areas such as medical image processing, biometric recognition, computer vision, 
optical character recognition, texture recognition or machine inspection of surfaces \cite{SonkaHlavacBoyle2008,StegerUlrichWiedemann2008,Davies2012}.
Methods for image enhancement can be grouped by the domain in which they perform their operations:
Images are processed in the spatial domain or Fourier domain,
or modified e.g. in the wavelet or curvelet domain \cite{MaPlonka2010}.
Types of enhancement methods include contextual filtering, 
e.g. for fingerprint image enhancement~\cite{Gottschlich2012,GottschlichSchoenlieb2012,BartunekNilssonSallbergClaesson2013},
contrast enhancement, e.g. by histogram equalization \cite{Marukatat2015},
and image superresolution \cite{YangWrightHuangMa2010}.
Image restoration is connected to the notion that a given input image
suffers from degradation and the goal is restore an ideal version of it.
Degradations are caused by various types of noise, missing pixels or blurring
and their countermeasures are denoising, inpainting and deblurring.
In general, one has to solve a linear or nonlinear inverse problem to reconstruct the ideal image 
from its given degraded version.
Denoising aims to remove noise from an image and denoising methods 
include total variation minimization based approaches \cite{VeseOsher2004},
the application of nonlocal means (NL-Means) \cite{BuadesCollMorel2005}
or other dictionaries of image patches for smoothing,
and adaptive thresholding in the Wavelet domain \cite{ChangYuVetterli2000}.
Inpainting \cite{Schoenlieb2009PhD} is the filling-in of missing pixels
from the available information in the image
and it is applied for scratch removal from scanned photographs, 
for occlusion filling, 
for removing objects or persons from images (in image forgery \cite{BattistiCarliNeri2012} 
or for special effects), for filling-in of pixels which were lost during the transmission
of an image or left out on purpose for image compression \cite{Salomon2007Compression}.
Deblurring \cite{JoshiKangZitnickSzeliski2010} addresses the removal of blurring artifacts 
and is not in the focus of this paper.

Rudin, Osher, and Fatemi \cite{RudinOsherFatemi1992} pioneered two-part image decomposition
by total variation (TV) regularization for denoising.
Shen and Chan \cite{ShenChan2002} applied TV regularization to image inpainting, called TV inpainting model.
and they also suggested image inpainting by curvature-driven diffusions (CDD), see \cite{ChanShen2001}. 
Starck et. al. \cite{StarckEladDonoho2005} defined a model for two-part decomposition based on dictionary approach.
Then, Elad \textit{et al.} \cite{EladStarckQuerreDonoho2005} applied this decomposition idea for image inpainting
by introducing the indicator function in the $\ell_2$ norm of the residual, 
see Eq. (6) in \cite{EladStarckQuerreDonoho2005}.
Esedoglu and Shen \cite{EsedogluShen2002}
introduced two inpainting models based on the Mumford-Shah model \cite{MumfordShah1989} 
and its higher order correction - the Mumford-Shah-Euler image model.
They also presented numerical computation based on the $\Gamma$-convergence approximations 
\cite{AmbrosioTortorelli1990, Giorgi1991}. 
Shen \textit{et al.} \cite{ShenKangChan2002} proposed image inpainting based on bounded variation 
and elastica models for non-textured images.

Image inpainting can be an easy or difficult problem depending on the amount of missing pixels \cite{ChanShen2001},
the complexity of the image content and whether prior knowledge about the image content is available.
Methods have been proposed which perform 
only cartoon inpainting (also referred to as structure inpainting) 
\cite{ShenChan2002,ShenKangChan2002,BallesterBertalmioCasellesSapiroVerdera2001}
or only texture inpainting \cite{EfrosLeung1999}.
Images which consist of both cartoon (structure) and texture components are more challenging to inpaint.
Bertalmio \textit{et al.} \cite{BertalmioVeseSapiroOsher2003}, 
Elad \textit{et al.} \cite{EladStarckQuerreDonoho2005}
and Cai \textit{et al.} \cite{CaiChanShen2010}
have proposed methods for inpainting 
which can handle images with both cartoon (structure) and texture components.

In this paper, we tackle an even more challenging problem. Consider
an input image~$\Bf$ which has the following three properties:
\begin{itemize}
\item[(i)]~a large percentage of pixels in $\Bf$ are missing and shall be inpainted.
\item[(ii)]~the known pixels in $\Bf$ are corrupted by noise.
\item[(iii)]~$\Bf$ contains both cartoon and texture elements.
\end{itemize}

The co-occurrence of noise and missing pixels in an image with cartoon and texture components
increases the difficulty of 
both the inpainting problem and the denoising problem. 
A multitude of methods has been proposed for inpainting and denoising.
Existing inpainting methods in the literature typically assume that the non-missing pixels 
in a given image contain only a small amount of noise or are noise-free,
and existing methods for denoising typically assume all pixels of the noisy image are known.
The proposed method for solving this challenging problem 
is inspired by the works of 
Efros and Leung~\cite{EfrosLeung1999},
Bertalmio \textit{et al.} \cite{BertalmioVeseSapiroOsher2003}, 
Vese and Osher \cite{VeseOsher2003},
Aujol and Chambolle \cite{AujolChambolle2005},
Buades \textit{et al.} \cite{BuadesCollMorel2005} and
Elad \textit{et al.} \cite{EladStarckQuerreDonoho2005},
and it is based on the directional global three-part decomposition (DG3PD) \cite{ThaiGottschlich2015DG3PD}.
The DG3PD method decomposes an image into three parts: a cartoon image, a texture image and a residual image.
Advantages of the DG3PD
model lie in the properties which are enforced on the cartoon and texture images.
The geometric objects in the cartoon image have a very smooth surface and sharp edges. 
The texture image yields oscillating patterns on a defined scale which is both smooth and sparse.
Recently, the texture images have been applied as a very useful feature 
for fingerprint segmentation \cite{ThaiGottschlich2015DG3PD,ThaiGottschlich2015G3PD,ThaiHuckemannGottschlich2015}.

\begin{figure}[ht] \label{fig:Overview}
\begin{center}    
  \includegraphics[width=0.9\textwidth]{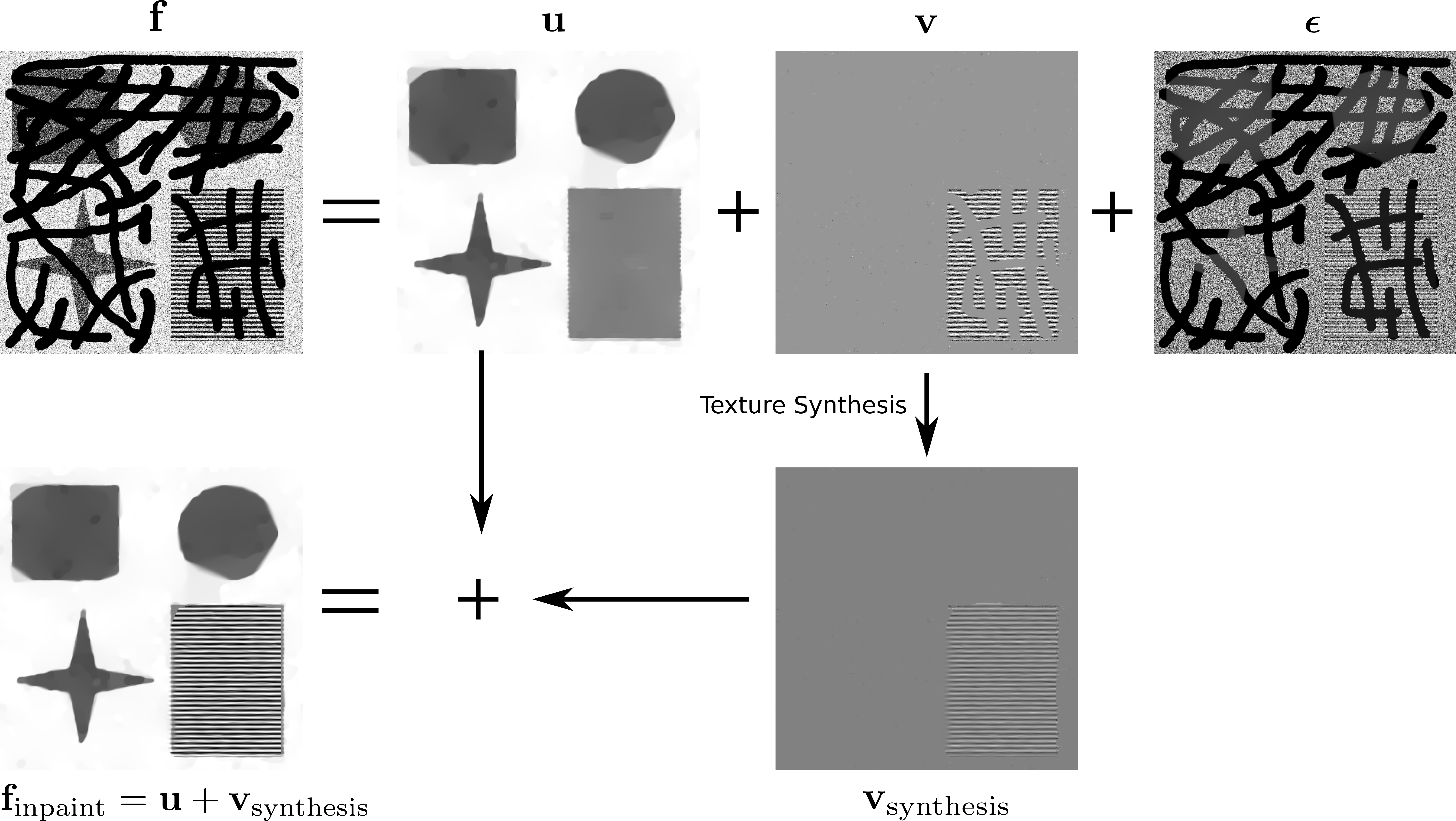}
  \caption{Overview over the DG3PD image inpainting and denoising process.}
\end{center}
\end{figure}

We address the challenging task of simultaneous inpainting and denoising in the following way:
The advanced DG3PD model introduced in the next section
decomposes a noisy input image $\Bf$ (with missing regions $D$)
into cartoon $\Bu$, texture $\Bv$ and residual $\Beps$ components.
At the same time, the missing regions $D$ of the cartoon component $\Bu$ are interpolated
and the available regions of $\Bu$ are denoised
by the advantage of multi-directional bounded variation.
This effect benefits from the help of the indicator function in the measurement of the residual, i.e. $\norm{ \cC \{ \chi_D^c \cdot^\times \Beps \} }_{\ell_\infty}$ in (1).
However, texture $\Bv$ is not interpolated due to the "cancelling" effect of this supremum norm for residual in unknown regions.
Therefore, the texture component $\Bv$ is inpainted and denoised by a dictionary based approach instead.
The DG3PD decomposition drives noise into the residual component $\Beps$ which is discarded.
The reconstruction of the ideal version of $\Bf$ is obtained by summation 
of the inpainted and denoised cartoon and texture components 
(see Figure \ref{fig:Overview} for an visual overview).

Moreover, we uncover the link between the calculus of variations \cite{AubertKornprobst2002,ScherzerBookVariationalMethods2009,Scherzer2011}
and filtering in the Fourier domain \cite{PrandoniVetterli2008} by analyzing the solution of 
the convex minimization in Eq. (\ref{eq:DG3PD_inpainting}), i.e. roughly speaking the solution of the DG3PD inpainting model 
which can be understood as
the response of the lowpass filter $\widehat{\text{LP}}(\Bome)$, highpass filter $\widehat{\text{HP}}(\Bome)$ and 
bandpass filter $\widehat{\text{BP}}(\Bome)$ and the unity condition is satisfied, i.e.
$$
\widehat{\text{LP}}(\Bome) + \widehat{\text{BP}}(\Bome) + \widehat{\text{HP}}(\Bome) = 1,
$$
where $\Bome \in [-\pi \,, \pi]^2$ is a coordinator in the Fourier domain. 
We observe that this decomposition is similar to wavelet 
or pyramidal decomposition scheme \cite{BurtAdelson1983,SimoncelliFreeman1995,UnserChenouardVandeville2011}.
However, the basis elements obtaining the decomposition, i.e. scaling function and frame (or wavelet-like) function,
are constructed 
by discrete differential operators (due to the discrete setting in minimizing (1))
which are referred to as wavelet-like operators in \cite{KhalidovUnser2006}.
In particular,
\begin{itemize}
 \item the scaling function and wavelet-like function for the cartoon $\Bu$ are from the effect of the multi-directional TV norm,
 \item the scaling function and wavelet-like function to extract the texture $\Bv$ are reconstructed by the effect of the multi-directional G-norm,
 \item the effect of $\ell_\infty$ norm $\norm{ \cC\{ \chi_D^c \cdot^\times \Beps\} }_{\ell_\infty}$
       is to remove the remaining signal in the known regions of the residual $\Beps$ (due to the duality property of $\ell_\infty$).
\end{itemize}
We also describe flowcharts to show that the method of variational calculus (or the DG3PD inpainting)
is a closed loop pyramidal decomposition which is different from an open loop one, 
e.g. wavelet \cite{Mallat2008}, curvelet \cite{CandesDonoho2004}, see Figure \ref{fig:VariationPyramid}.
By numerics, we observe that the closed loop filter design by the calculus of variation will result in lowpass, 
highpass and bandpass filters which are ``unique'' for different images, see Figure \ref{fig:Fourier:uniqueFilter}.
We also analyze the DG3PD inpainting model from a view of the Bayesian framework 
and then define a discrete innovation model for this inverse problem.

The setup of paper is as follows.
In Section \ref{sec:InpaintingDG3PD}, we describe the DG3PD model for image inpainting and denoising.
In Section \ref{sec:solutionDG3PD}, we show how to compute the solution of the convex minimization in
the DG3PD inpainting problem by augmented Lagrangian method.
In Section \ref{sec:textureInpainting}, we describe the proposed method for texture inpainting and denoising.
In Section \ref{sec:comparisonPriorArt}, we compare the proposed method 
to existing ones (TVL2 inpainting, Telea \cite{Telea2004} and Navier Stokes \cite{BertalmioBertozziSapiro2001}).
In Section \ref{sec:BayesianDecomposition}, we consider our inverse problem from a statistical point of view, i.e.
the Bayesian framework, to describe how to select priors for cartoon $\Bu$ and texture $\Bv$.
We analyze the relation between the calculus of variations and the traditional pyramid decomposition scheme, 
e.g. Gaussian pyramid, in Section \ref{sec:VariationalFourierConnection}. 
We conclude the study with Section \ref{sec:conclusion}.
For more detailed notation and mathematical preliminaries, we refer the reader to \cite{ThaiGottschlich2015G3PD,Thai2015PhD}.

\section{Inpainting by DG3PD} \label{sec:InpaintingDG3PD}

We define a method for a restoration of the original noisy/non-noisy image $\Bf$ with a set of known missing regions $D$.
The proposed model is a generalized version of DG3PD \cite{ThaiGottschlich2015DG3PD} for the inpainting and denoising problem.
This modification for our DG3PD-inpainting is inspired by \cite{EladStarckQuerreDonoho2005}.

In particular, the discrete image $\Bf$ (size $m \times n$) with missing regions $D$ 
is simultaneously decomposed into cartoon $\Bu$, texture $\Bv$ and noise $\Beps$,
especially, cartoon $\Bu$ is interpolated on the missing regions due to the indicator function $\chi_D$ in the modified model.

A set of missing regions $D$ on a bounded domain $\Omega$ is defined by the indicator function $\chi_D$
whose complement is 
$$
\chi_D^c[\Bk] = \begin{cases} 1 \,, & \Bk \in \Omega \backslash D  \\  0 \,, & \Bk \in D \end{cases}.
$$

Instead of putting $\chi_D^c$ on $\ell_2$ norm of the residual (see Eq. (4) and (6) in \cite{EladStarckQuerreDonoho2005}),
we introduce $\chi_D^c$ for the residual in the DG3PD model with the point-wise multiplication operator $\cdot^\times$, i.e. 
$\norm{\mathcal C\big\{ \chi_D^c \cdot^\times \Beps \big\}}_{\ell_\infty}$.
We propose the DG3PD-inpainting as follows
\begin{align} \label{eq:DG3PD_inpainting}
 &(\Bu^* \,, \Bv^* \,, \Beps^* \,, \Bg^*) = \argmin_{(\Bu, \Bv, \Beps, \Bg) \in X^{3+S}} \bigg\{ 
 \underbrace{ \sum_{l=0}^{L-1} \norm{\cos\big(\frac{\pi l}{L}\big) \Bu \BDnT + \sin\big(\frac{\pi l}{L}\big) \BDm \Bu }_{\ell_1} }
 _{ :=~ \norm{\nabla^+_L \Bu}_{\ell_1} }
 + \mu_1 \underbrace{ \sum_{s=0}^{S-1} \norm{\Bg_s}_{\ell_1} }_{:=~ \norm{\Bv}_{G_S}}
 + \mu_2 \norm{\Bv}_{\ell_1}       \notag
 \\
 & \text{s.t.  }
 \norm{\mathcal C\big\{ \chi_D^c \cdot^\times \Beps \big\}}_{\ell_\infty} \leq \nu \,,~
 \Bv = \underbrace{ \sum_{s=0}^{S-1} \Big[ \cos\big(\frac{\pi s}{S}\big) \Bg_s \BDnT + \sin\big(\frac{\pi s}{S}\big) \BDm \Bg_s \Big] }
                 _{ = \text{div}^+_S \Bg } \,,~
 \Bf = \Bu + \Bv + \Beps
 \bigg\}.
\end{align}

Note that if there are no missing regions, i.e. $\chi_D^c[\Bk] = 1 \,, \forall \Bk \in \Omega$, 
the minimization problem (\ref{eq:DG3PD_inpainting}) becomes the DG3PD model \cite{ThaiGottschlich2015DG3PD}.
Next, we discuss how to solve the DG3PD-inpainting model (\ref{eq:DG3PD_inpainting}).

\section{Solution of the DG3PD Inpainting Model} \label{sec:solutionDG3PD}

In this section we present a numerical algorithm
for obtaining the solution of the DG3PD-inpainting model stated in (\ref{eq:DG3PD_inpainting}).
To simplify the calculation, we introduce three new variables:
\begin{equation*}
 \begin{cases}
  \Br_b = \cos\big( \frac{\pi b}{L} \big) \Bu \BDnT + \sin\big( \frac{\pi b}{L} \big) \BDm \Bu \,,~ b = 0 \,, \ldots \,, L-1 \,, \\
  \Bw_a = \Bg_a \,, a = 0 \,, \ldots \,, S-1 \,, \\
  \Be = \chi_D^c \cdot^\times \Beps \,.
 \end{cases}
\end{equation*}
and denote $G^*\big( \frac{\Be}{\nu} \big)$ as the indicator function on 
the feasible convex set $A(\nu)$ of (\ref{eq:DG3PD_inpainting}), i.e.
\begin{equation*}
 A(\nu) ~=~ \Big\{ \Be \in X \,:~ \norm{C \big\{ \Be \big\}}_{\ell_\infty} \leq \nu \Big\} 
 \text{  and  }
 G^*\big( \frac{\Be}{\nu} \big) ~=~ \begin{cases} 0 \,, &\Beps \in A(\nu)  \\  +\infty \,, &\text{else} \end{cases}.
\end{equation*}
Due to the constrained minimization problem in (\ref{eq:DG3PD_inpainting}), 
the augmented Lagrangian method is applied to turn it into an unconstrained one as
\begin{equation} \label{eq:DG3PD_inpainting_ALM}
 \min_{\big( \Bu \,, \Bv \,, \Beps \,, \Be \,, \big[ \Br_l \big]_{l=0}^{L-1} \,,
 \big[ \Bw_s \big]_{s=0}^{S-1} \,, \big[ \Bg_s \big]_{s=0}^{S-1} \big) \in X^{L+2S+4} }
 \mathcal L \Big( \cdot \;; \big[ \boldsymbol{\lambda}_{\boldsymbol{1} l} \big]_{l=0}^{L-1} \,,
 \big[ \boldsymbol{\lambda}_{\boldsymbol{2}s} \big]_{s=0}^{S-1} \,, \boldsymbol{\lambda_3} \,, \boldsymbol{\lambda_4} \,, \boldsymbol{\lambda_5} \Big) \,,
\end{equation}
where the Lagrange function is
\begin{align*}
 \mathcal L ( \cdot \;; \cdot ) &=
 \sum_{l=0}^{L-1} \norm{ \Br_l }_{\ell_1} 
 + \mu_1 \sum_{s=0}^{S-1} \norm{\Bw_s}_{\ell_1} + \mu_2 \norm{\Bv}_{\ell_1} + G^* \big( \frac{\Be}{\nu} \big)
 \\&
 + \frac{\beta_1}{2} \sum_{l=0}^{L-1} \norm{ \Br_l - \cos\big( \frac{\pi l}{L} \big) \Bu \BDnT - \sin\big( \frac{\pi l}{L} \big) \BDm \Bu 
 + \frac{\boldsymbol{\lambda}_{\boldsymbol{1} l}}{\beta_1} }^2_{\ell_2}
 \\&
 + \frac{\beta_2}{2} \sum_{s=0}^{S-1} \norm{\Bw_s - \Bg_s + \frac{\boldsymbol{\lambda}_{\boldsymbol{2}s}}{\beta_2}}^2_{\ell_2}
 + \frac{\beta_3}{2} \norm{ \Bv - \sum_{s=0}^{S-1} \Big[ \cos\big( \frac{\pi s}{S} \big) \Bg_s \BDnT 
 + \sin\big( \frac{\pi s}{S} \big) \BDm \Bg_s \Big] + \frac{\boldsymbol{\lambda_3}}{\beta_3} }^2_{\ell_2}
 \\&
 + \frac{\beta_4}{2} \norm{ \Bf - \Bu - \Bv - \Beps + \frac{\boldsymbol{\lambda_4}}{\beta_4} }_{\ell_2}^2
 + \frac{\beta_5}{2} \norm{ \Be - \chi_D^c \cdot^\times \Beps + \frac{\boldsymbol{\lambda_5}}{\beta_5} }_{\ell_2}^2
 \Bigg\} .
\end{align*}
Similar to \cite{ThaiGottschlich2015DG3PD}, 
the alternating directional method of multipliers is applied to solve the unconstrained minimization problem 
(\ref{eq:DG3PD_inpainting_ALM}).
Its minimizer is numerically computed through iterations $t = 1 \,, 2 \,, \ldots$
\begin{align} \label{eq:DG3PD_inpainting_ALM_numerical} 
 &{\Big( \Bu^{(t)} \,, \Bv^{(t)} \,, \Beps^{(t)} \,, \Be^{(t)} \,, \big[ \Br_l^{(t)} \big]_{l=0}^{L-1} \,,
 \big[ \Bw_s^{(t)} \big]_{s=0}^{S-1} \,, \big[ \Bg_s^{(t)} \big]_{s=0}^{S-1} \Big) }                     \notag
 ~=~
 \\
 &\argmin ~ \mathcal L \Big( \Bu \,, \Bv \,, \Beps \,, \Be \,, \big[ \Br_l \big]_{l=0}^{L-1} \,,
 \big[ \Bw_s \big]_{s=0}^{S-1} \,, \big[ \Bg_s \big]_{s=0}^{S-1} \;;~ \big[ \boldsymbol{\lambda}_{\boldsymbol{1} l}^{(t-1)} \big]_{l=0}^{L-1} \,,
 \big[ \boldsymbol{\lambda}_{\boldsymbol{2}s}^{(t-1)} \big]_{s=0}^{S-1} \,, \boldsymbol{\lambda}_{\mathbf 3}^{(t-1)} \,, 
 \boldsymbol{\lambda}_{\mathbf 4}^{(t-1)} \,, \boldsymbol{\lambda}_{\mathbf 5}^{(t-1)} \Big)
\end{align}
and the Lagrange multipliers are updated after every step $t$.
We also initialize 
$\Bu^{(0)} = \Bf \,, \Bv^{(0)} = \Beps^{(0)} = \Be^{(0)} = \big[ \Br_l^{(0)} \big]_{l=0}^{L-1} 
= \big[ \Bw_s^{(0)} \big]_{s=0}^{S-1} = \big[ \Bg_s^{(0)} \big]_{s=0}^{S-1} 
= \big[\boldsymbol{\lambda}_{\mathbf{1}l}^{(0)}\big]_{l=0}^{L-1} = \big[\boldsymbol{\lambda}_{\mathbf{2}a}^{(0)}\big]_{a=0}^{S-1} 
= \boldsymbol{\lambda}_{\mathbf 3}^{(0)} = \boldsymbol{\lambda}_{\mathbf 4}^{(0)} = \boldsymbol{\lambda}_{\mathbf 5}^{(0)} =\boldsymbol 0$,
where $\boldsymbol 0$ is a $m \times n$ zero matrix.

In each iteration, we first solve the seven subproblems in the listed order:
``$\big[ \Br_l \big]_{l=0}^{L-1}$-problem'', ``$\big[ \Bw_s \big]_{s=0}^{S-1}$-problem'',
``$\big[ \Bg_s \big]_{s=0}^{S-1}$-problem'', ``$\Bv$-problem'', ``$\Bu$-problem'',
``$\Be$-problem'', ``$\Beps$-problem'', 
and then we update the five Lagrange multipliers, namely 
$\big[\boldsymbol{\lambda}_{\mathbf{1l}}\big]_{l=0}^{L-1} \,, \big[\boldsymbol{\lambda}_{\mathbf{2a}}\big]_{a=0}^{S-1} 
\,, \boldsymbol{\lambda_3}\,,  \boldsymbol{\lambda_4} \,, \boldsymbol{\lambda_5}$, see Algorithm 1.

In this section, we only present the solution of the ``$\Be$-problem'' and ``$\Beps$-problem''
as well as the updated Lagrange multiplier $\boldsymbol{\lambda_5}$.
We refer the readers to \cite{ThaiGottschlich2015DG3PD} for the detailed explanation of
the other subproblems and the other Lagrange multipliers.

{\bfseries The $\Be$-problem:}
Fix $\Bu$, $\Bv$, $\Beps$, $\big[ \Br_l \big]_{l=0}^{L-1}$, $\big[ \Bw_s \big]_{s=0}^{S-1}$, $\big[ \Bg_s \big]_{s=0}^{S-1}$ and
\begin{equation} \label{eq:sub:e}
 \min_{\Be \in X} \Big\{ 
 G^* \big( \frac{\Be}{\nu} \big) + 
 \frac{\beta_5}{2} \norm{ \Be - \Big( \chi_D^c \cdot^\times \Beps - \frac{\boldsymbol{\lambda_5}}{\beta_5} \Big) }_{\ell_2}^2
 \Big\}
\end{equation}
The solution of (\ref{eq:sub:e}) is defined as a projection operator on a convex set $A(\nu)$,
i.e. $\mathbb P_{A(\nu)} \Big( \chi_D^c \cdot^\times \Beps - \frac{\boldsymbol{\lambda_5}}{\beta_5} \Big)$:
\begin{equation} \label{eq:sub:e:solution}
 \Be^* = \Big( \chi_D^c \cdot^\times \Beps - \frac{\boldsymbol{\lambda_5}}{\beta_5} \Big) ~-~
 \CST \Big( \chi_D^c \cdot^\times \Beps - \frac{\boldsymbol{\lambda_5}}{\beta_5} \,, \nu \Big)
\end{equation}

{\bfseries The $\Beps$-problem:}
Fix $\Bu$, $\Bv$, $\Be$, $\big[ \Br_l \big]_{l=0}^{L-1}$, $\big[ \Bw_s \big]_{s=0}^{S-1}$, $\big[ \Bg_s \big]_{s=0}^{S-1}$ and
\begin{equation} \label{eq:sub:epsilon}
 \min_{\Beps \in X} \Big\{ 
 \frac{\beta_4}{2} \norm{ \Bf - \Bu - \Bv - \Beps + \frac{\boldsymbol{\lambda_4}}{\beta_4} }_{\ell_2}^2
 + \frac{\beta_5}{2} \norm{ \Be - \chi_D^c \cdot^\times \Beps + \frac{\boldsymbol{\lambda_5}}{\beta_5} }_{\ell_2}^2 
 \Big\}
\end{equation}
The solution of (\ref{eq:sub:epsilon}) with the point-wise division operator $/_.$ is
\begin{equation} \label{eq:sub:epsilon:solution}
 \Beps^* = \Big[ \beta_4 \big( \Bf - \Bu - \Bv + \frac{\boldsymbol{\lambda_4}}{\beta_4} \big)
 + \beta_5 \chi^c_D \cdot^\times \big( \Be + \frac{\boldsymbol{\lambda_5}}{\beta_5} \big) \Big]
 /_. \Big[ \beta_4 \mathbf{1_{mn}} + \beta_5 \chi^c_D \Big]
\end{equation}
and $\mathbf{1_{mn}}$ is a $m \times n$ matrix of ones.

{\bfseries Updated Lagrange Multiplier $\boldsymbol{\lambda_5} \in X$:}
\begin{equation*}
 \boldsymbol{\lambda_5} = \boldsymbol{\lambda_5} + \beta_5 \Big( \Be - \chi^c_D \cdot^\times \Beps \Big)
\end{equation*}

{\bfseries Choice of Parameters}

Due to an adaptability to specific images and a balance between the smoothing terms and updated terms for the solution of the above subproblems,
the selection of parameters $(\mu_1 \,, \mu_2 \,, \beta_1 \,, \beta_2 \,, \beta_3)$ is described in \cite{ThaiGottschlich2015DG3PD}
and $\beta_5$ is defined as
\begin{equation*}
 \beta_5 = c_3 \beta_4.
\end{equation*}
\begin{figure}
\begin{center}    
  \includegraphics[width=1\textwidth]{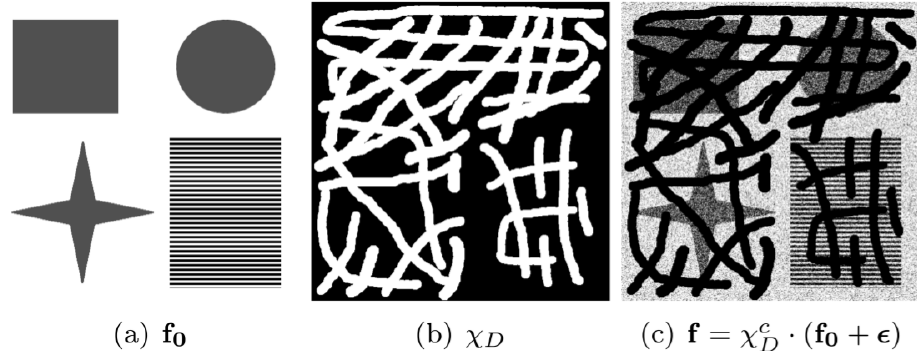}
  
  \caption{An ideal image (a) with cartoon and texture components. Missing pixels (b) are shown in white.
	         Image (c) combines missing pixels and noise.
	         $\epsilon[\Bk] \sim \mathcal N(0 \,, \sigma^2 = 100^2) \,, \Bk \in \Omega$. }
\label{fig:challenge}
\end{center}
\end{figure}

\begin{figure}[ht]
\begin{center}    
  \includegraphics[width=1\textwidth]{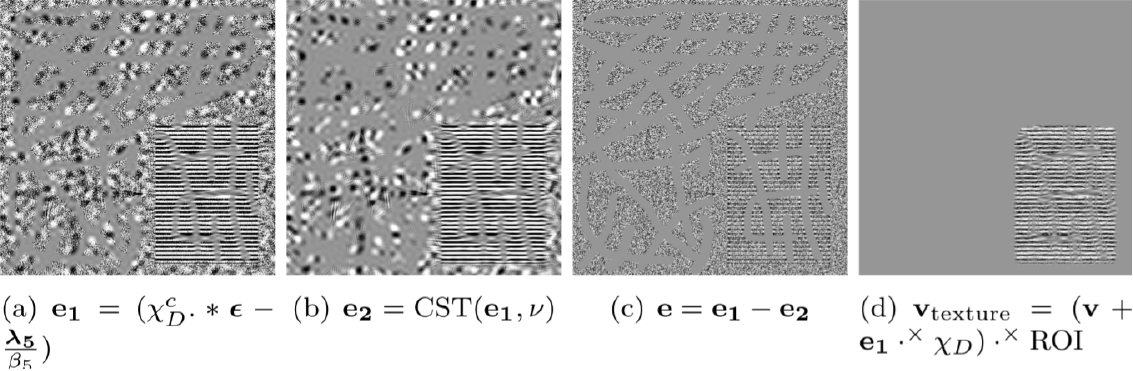}
  
  \caption{The ``canceling effect'' is caused by the noise measurement $\norm{\cC \{ \chi_D^c \cdot^\times \Beps \}}_{\ell_\infty}$.}
\label{fig:DG3PDinpainting:textureDG3PD}
\end{center}
\end{figure}

The term $\norm{\cC \{ \chi_D^c \cdot^\times \Beps \}}_{\ell_\infty}$ in (\ref{eq:DG3PD_inpainting}) serves as a good measurement for the amount of noise
due to the substraction in Eq. (\ref{eq:sub:e:solution}), see \cite{CandesDemanetDonohoYing2006, ThaiGottschlich2015G3PD, ThaiGottschlich2015DG3PD}
(thanks to the multiscale and multidirection properties of the curvelet transform and its duality).
However, due to the indicator function $\chi_D^c$ in the noise measurement 
$\norm{\cC \{ \chi_D^c \cdot^\times \Beps \}}_{\ell_\infty}$, the interpolated texture by DG3PD inpainting 
is almost in the residual, see Figure \ref{fig:DG3PDinpainting:textureDG3PD} (a), and its smoothed version by curvelet shrinkage (b).
Because of the substraction operator in Eq. (\ref{eq:sub:e:solution}), these interpolated textures are canceled out in $\Be$, see (c).
Figure \ref{fig:DG3PDinpainting:textureDG3PD} (d) illustrates the estimated texture before the 
substraction in Eq. (\ref{eq:sub:e:solution}) as
$$\Bv_\text{texture} = (\Bv + \mathbf{e_1} \cdot^\times \chi_D)\cdot^\times \text{ROI} \,,$$
where ROI is the region of interest obtained by the morphological operator for texture image in \cite{ThaiHuckemannGottschlich2015}.

\begin{figure}
\begin{center}    
  \includegraphics[width=1\textwidth]{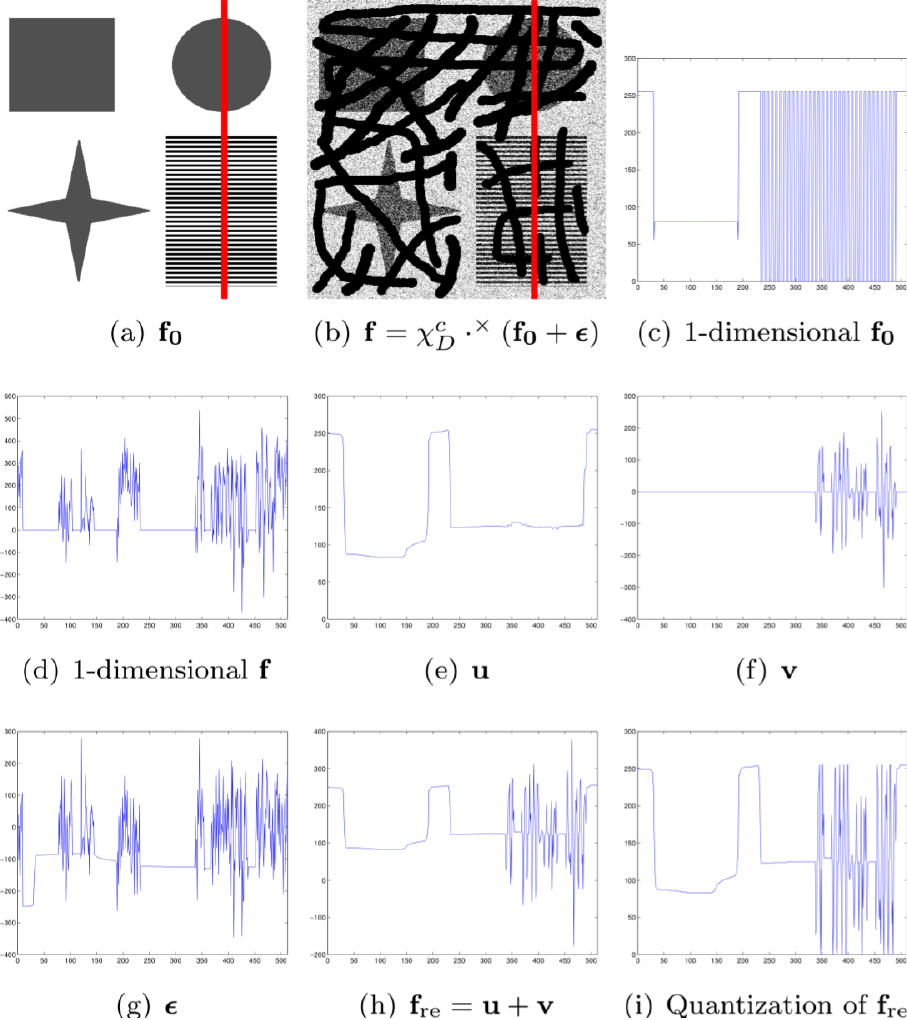}
  
  \caption{Interpolation for 1-dimensional signal after DG3PD inpainting (without texture inpainting):
           (c) is an 1-dimensional signal extracted along the red line of the original image (a).
           (d) is its corrupted signal by Gaussian noise and a missing regions $D$, see (b).
           Thanks to the directional TV norm, the DG3PD inpainting produces ``sharp'' edges without ``stair-case'' effect in a cartoon $\Bu$ (e)
           and the mean values of $\Bu$ ``nearly'' remains.
           Due to noise and the shrinkage soft-thresholding operator by $\norm{\Bv}_{\ell_1}$ in (\ref{eq:DG3PD_inpainting}),
           texture $\Bv$ is reconstructed with a ``sharp'' transition on areas where texture presents in the original signal (c).
           However, due to a ``heavy'' noise $\Beps$ on the known domains in (e), the DG3PD fails to reconstruct
           a ``full'' texture. In general, this is a challenging problem to recover an oscillating (or weak) signal corrupted by heavy noise.
           (h) is a reconstructed image (without step of inpainting texture) and (i) is a quantized signal, 
           i.e. truncate to [0, 255] and quantize.
           }
\label{fig:DG3PD:Analysis}
\end{center}
\end{figure}

In order to analyze the effects of the proposed model in terms of interpolation (for $\Bu$) and decomposition,
we consider a 1-dimensional signal (which is extracted along the red line in Figure \ref{fig:DG3PD:Analysis} (a)-(d)).
By the DG3PD inpainting, the mean values of the cartoon $\Bu$ in (e) ``nearly'' remain unchanged.
The homogeneous regions have ``sharp'' edges and a ``stair-case'' effect does not occur (thanks to the directional TV).
Texture $\Bv$ in (f) is extracted in areas which contain repeated pattern in (b). 
Moreover, small scales objects, e.g. noise, are removed from $\Bv$. 

However, the term $\norm{\cC \{ \chi_D^c \cdot^\times \Beps \}}_{\ell_\infty}$ causes 
a ``canceling effect'' which removes the interpolated texture in unknown area during the decomposition process.
Therefore, texture inpainting and denoised is tackled separately as described in the next section 
by a generalized version of the dictionary learning for texture synthesis \cite{EfrosLeung1999}.

\begin{algorithm}
\label{alg:DG3PD}
\caption{The Discrete DG3PD Inpainting Model}
\begin{algorithmic}
\small
 \STATE{
  {\bfseries Initialization:}
  $\Bu^{(0)} = \Bf \,, \Bv^{(0)} = \Beps^{(0)} = \big[ \Br_l^{(0)} \big]_{l=0}^{L-1} = \big[ \Bw_s^{(0)} \big]_{s=0}^{S-1} = \big[ \Bg_s^{(0)} \big]_{s=0}^{S-1}
   = \big[\boldsymbol{\lambda}_{\mathbf{1}l}^{(0)}\big]_{l=0}^{L-1} = \big[\boldsymbol{\lambda}_{\mathbf{2}a}^{(0)}\big]_{a=0}^{S-1}
   = \boldsymbol{\lambda}_{\mathbf 3}^{(0)} = \boldsymbol{\lambda}_{\mathbf 4}^{(0)} =\boldsymbol 0$.
 } 
 \STATE{ } 
 \FOR{$t = 1 \,, \ldots \,, T $}
 \STATE
{
  {\bfseries 1. Compute}
   $\Big( \big[ \Br_b^{(t)} \big]_{b=0}^{L-1} \,, \big[ \mathbf{w}_a^{(t)} \big]_{a=0}^{S-1} \,,
    \big[ \mathbf{g}_a^{(t)} \big]_{a=0}^{S-1} \,, \Bv^{(t)} \,, \Bu^{(t)} \,, \Be^{(t)} \,, \Beps^{(t)} \Big) \in X^{L+2S+4}
   $:
   \begin{align*}
    \Br_b^{(t)} &~=~ \Shrink \Big( \cos\big( \frac{\pi b}{L} \big) \Bu^{(t-1)} \BDnT + \sin\big( \frac{\pi b}{L} \big) \BDm \Bu^{(t-1)}
    - \frac{\boldsymbol{\lambda}_{\boldsymbol{1} b}^{(t-1)}}{\beta_1} \,, \frac{1}{\beta_1} \Big)
    \,,~ \quad b = 0 \,, \ldots \,, L-1
    \\
    \mathbf{w}_a^{(t)} &~=~
    \Shrink \Big( \mathbf{t}_{\mathbf{w}_a} ~:=~ \mathbf{g}_a^{(t-1)} -
    \frac{ \boldsymbol{\lambda}_{\boldsymbol{2}a}^{(t-1)} }{\beta_2} \,, \frac{\mu_1}{\beta_2} \Big)
    \,,~ \quad a = 0 \,, \ldots \,, S-1
    \\
    \mathbf{g}_a^{(t)} &~=~ \RE \bigg[ \mathcal F^{-1} \Big\{ \frac{ \mathcal B_a^{(t)}(\Bz) }{ \mathcal A_a^{(t)}(\Bz) } \Big\} \bigg]
    \,,~ a = 0 \,, \ldots \,, S-1
    \\
    \Bv^{(t)} &= \Shrink \bigg( \mathbf{t_v} ~:=~ \frac{\beta_3}{\beta_3 + \beta_4}
    \bigg( \sum_{s=0}^{S-1} \Big[ \cos\big( \frac{\pi s}{S} \big) \Bg_s^{(t)} \BDnT
    + \sin\big( \frac{\pi s}{S} \big) \BDm \Bg_s^{(t)} \Big] - \frac{\boldsymbol{\lambda}_{\boldsymbol 3}^{(t-1)}}{\beta_3} \bigg)
    \\& \qquad \qquad \qquad \qquad
    + \frac{\beta_4}{\beta_3 + \beta_4} \bigg( \Bf - \Bu^{(t-1)} - \Beps^{(t-1)} + \frac{\boldsymbol{\lambda}_{\boldsymbol 4}^{(t-1)}}{\beta_4} \bigg)
    \,,~ \frac{\mu_2}{\beta_3 + \beta_4}
    \bigg)   
    \\
    \Bu^{(t)} &~=~ \RE \bigg[ \mathcal F^{-1} \Big\{ \frac{ \mathcal Y^{(t)}(\Bz) }{ \mathcal X^{(t)}(\Bz) } \Big\} \bigg]
    \\
    \Be^{(t)} &~=~ \Big( \chi_D^c \cdot^\times \Beps^{(t-1)} - \frac{\boldsymbol{\lambda}_{\boldsymbol 5}^{(t-1)}}{\beta_5} \Big) ~-~
    \CST \Big( \chi_D^c \cdot^\times \Beps^{(t-1)} - \frac{\boldsymbol{\lambda}_{\boldsymbol 5}^{(t-1)}}{\beta_5} \,, \nu \Big) \,, \quad
    \nu = c_\nu \norm{ \cC \{ \chi_D^c \cdot^\times \Beps^{(t-1)} - \frac{\boldsymbol{\lambda}_{\boldsymbol 5}^{(t-1)}}{\beta_5} \} }_{\ell_\infty} 
    \\
    \Beps^{(t)} &~=~ \Big[ \beta_4 \big( \Bf - \Bu^{(t)} - \Bv^{(t)} + \frac{\boldsymbol{\lambda}_{\boldsymbol 4}^{(t-1)}}{\beta_4} \big)
    + \beta_5 \chi^c_D \cdot^\times \big( \Be^{(t)} + \frac{\boldsymbol{\lambda}_{\boldsymbol 5}^{(t-1)}}{\beta_5} \big) \Big]
    /_. \Big[ \beta_4 \mathbf{1_{mn}} + \beta_5 \chi^c_D \Big]
   \end{align*} 
  {\bfseries 2. Update}
  $\Big( \big[\boldsymbol{\lambda}_{\mathbf{1}b}^{(t)}\big]_{b=0}^{L-1} \,, \big[\boldsymbol{\lambda}_{\mathbf{2}a}^{(t)}\big]_{a=0}^{S-1}
  \,, \boldsymbol{\lambda}_{\boldsymbol 3}^{(t)} \,, \boldsymbol{\lambda}_{\boldsymbol 4}^{(t)} \,, \boldsymbol{\lambda}_{\boldsymbol 5}^{(t)} \Big) \in X^{L+S+3}$:
  \begin{align*}
   \boldsymbol{\lambda}_{\mathbf{1}b}^{(t)} &~=~ \boldsymbol{\lambda}_{\mathbf{1}b}^{(t-1)}
   ~+~ \gamma \beta_1 \Big( \mathbf{r}_b^{(t)} - \cos\big( \frac{\pi b}{L} \big) \Bu^{(t)} \BDnT - \sin\big( \frac{\pi b}{L} \big) \BDm \Bu^{(t)}  \Big) 
   \,, \quad b = 0 \,, \ldots \,, L-1
   \\
   \boldsymbol{\lambda}_{\mathbf{2} a}^{(t)} &~=~ \boldsymbol{\lambda}_{\mathbf{2} a}^{(t-1)}
   ~+~ \gamma \beta_2 \Big( \mathbf{w}_a^{(t)} - \mathbf{g}_a^{(t)} \Big) 
   \,, \quad a = 0 \,, \ldots \,, S-1
   \\
   \boldsymbol{\lambda}_{\mathbf{3}}^{(t)} &~=~ \boldsymbol{\lambda}_{\mathbf{3}}^{(t-1)}
   ~+~ \gamma \beta_3 \Big( \Bv^{(t)} - \sum_{s=0}^{S-1} \big[ \cos\big(\frac{\pi s}{S} \big) \mathbf{g}_s^{(t)} \BDnT
   + \sin\big( \frac{\pi s}{S} \big) \BDm \mathbf{g}_s^{(t)} \big] \Big) 
   \\
   \boldsymbol{\lambda}_{\mathbf{4}}^{(t)} &~=~ \boldsymbol{\lambda}_{\mathbf{4}}^{(t-1)}
   ~+~ \gamma \beta_4 \big( \Bf - \Bu^{(t)} - \Bv^{(t)} - \Beps^{(t)} \big) 
   \\
   \boldsymbol{\lambda}_{\boldsymbol5}^{(t)} &~=~ \boldsymbol{\lambda}_{\boldsymbol 5}^{(t-1)} + \beta_5 \Big( \Be^{(t)} - \chi^c_D \cdot^\times \Beps^{(t)} \Big)
  \end{align*}
 }
\ENDFOR
\end{algorithmic}
\end{algorithm}

\begin{algorithm}
\label{alg:DG3PD_Part2}
\begin{algorithmic}
\small
\STATE
\begin{align*}
 &\mathcal A_a(\Bz) ~=~ \beta_2 \mathbf{1_{mn}} + \beta_3 \abs{ \sin \frac{\pi a}{S} (z_1 - 1) + \cos \frac{\pi a}{S} (z_2 - 1) }^2
 \,,
 \\
 &\mathcal B_a(\Bz) ~=~
 \beta_2 \Big[ W_a(\Bz)  + \frac{\Lambda_{2a}(\Bz) }{\beta_2} \Big]
 ~+~ \beta_3 \Big[ \sin\big( \frac{\pi a}{S} \big) (z_1^{-1} -1) + \cos\big( \frac{\pi a}{S} \big) (z_2^{-1}-1) \Big] \times
 \\&
 \bigg[ V(\Bz) - \sum_{s=[0\,,S-1] \backslash \{a\} }
 \Big[ \cos\big( \frac{\pi s}{S} \big) (z_2-1)
 + \sin\big( \frac{\pi s}{S} \big) (z_1-1) \Big] G_s(\Bz)
 + \frac{\Lambda_3(\Bz)}{\beta_3} \bigg] \,,
 \\
 &\mathcal X(\Bz) =
 \beta_4 \mathbf{1_{mn}} + \beta_1 \sum_{l=0}^{L-1} \abs{ \sin\big( \frac{\pi l}{L} \big) (z_1-1) + \cos\big( \frac{\pi l}{L} \big) (z_2-1) }^2 \,,
 \\
 &\mathcal Y(\Bz) =
 \beta_4 \Big[ F(\Bz) - V(\Bz) - \mathcal{E}(\Bz) + \frac{\Lambda_4(\Bz)}{\beta_4} \Big]
 + \beta_1 \sum_{l=0}^{L-1} \Big[ \sin \big( \frac{\pi l}{L} \big) (z_1^{-1}-1) + \cos \big( \frac{\pi l}{L} \big) (z_2^{-1}-1) \Big]
 \Big[ R_l(\Bz) + \frac{\Lambda_{1l}(\Bz)}{\beta_1} \Big] .
\end{align*}
{\bfseries Choice of Parameters}
\begin{align*}
 \mu_1 = c_{\mu_1} \beta_2 \cdot \max_{\Bk \in \Omega} \big( \abs{t_{\Bw_a}[\Bk]} \big) \,,
 \mu_2 = c_{\mu_2} (\beta_3 + \beta_4) \cdot \max_{\Bk \in \Omega} \big( \abs{t_\Bv[\Bk]} \big) \,,
 \beta_2 = c_2 \beta_3 \,, \beta_3 = \frac{\theta}{1 - \theta} \beta_4 \,, \beta_1 = c_1 \beta_4 \,, \beta_5 = c_3 \beta_4.
\end{align*}
\end{algorithmic}
\end{algorithm}


\section{Texture Inpainting and Denoising} \label{sec:textureInpainting}

\begin{figure}[th] 
\begin{center}    
  \includegraphics[width=1\textwidth]{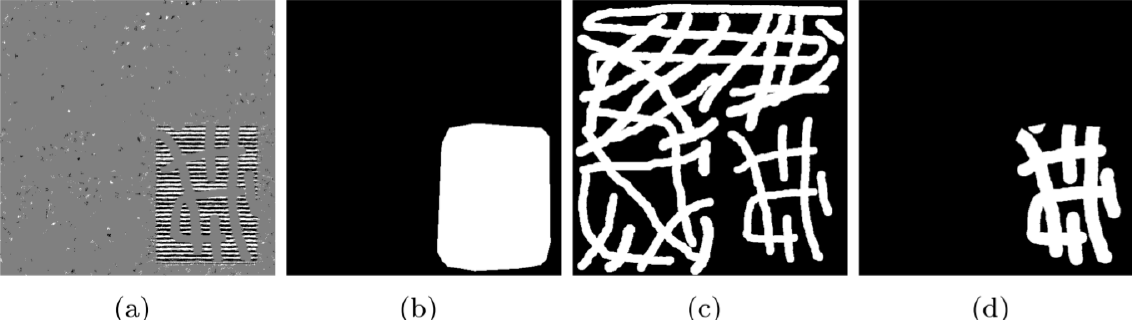}

  \caption{All coefficients of the obtained texture component $\Bv$ equal to zero are visualized by gray pixels in (a),
	         positive coefficients are indicated by white pixels and negative coefficients by black pixels.
					 (b) depicts the resulting segmentation. The mask of missing pixels $D$ shown in (c) is dilated 
					 and then intersected with (b), leading to the (white) pixels which are to be inpainted with texture in (d). 
					 \label{fig:TextureMorphology}}
\end{center}
\end{figure}

\begin{figure}[th] 
\begin{center}    
  \includegraphics[width=1\textwidth]{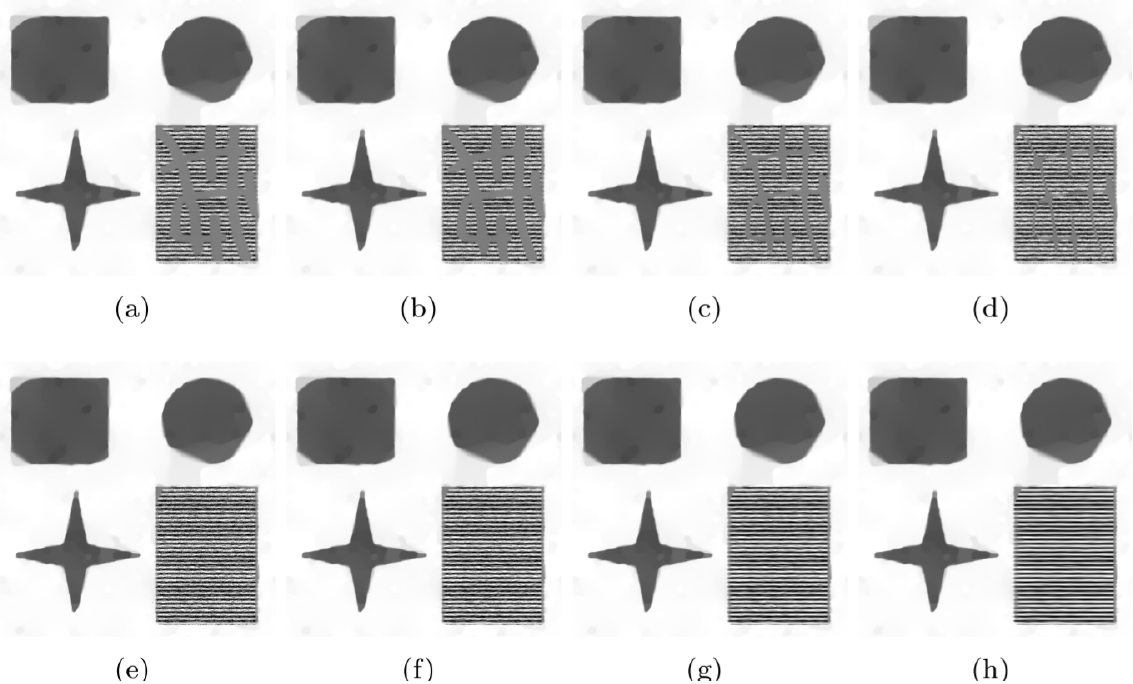}

  \caption{Images (a) to (e) visualize the inpainting progress (added to the cartoon component) 
	         from 0 to 100\% in steps of 25\%. 
					 Images (f) to (h) display the denoising results after one, five and 20 iterations, respectively.
					 \label{fig:TextureInpaintingAndDenoising}}
\end{center}
\end{figure}

The proposed method for reconstructing the texture component combines the following ideas 
in five subsequent steps. 

\begin{itemize}
\item Texture extraction by DG3PD \cite{ThaiGottschlich2015DG3PD}.
\item Morphological processing \cite{ThaiGottschlich2015G3PD,ThaiHuckemannGottschlich2015}.
\item Texture inpainting inspired by the work of Efros and Leung \cite{EfrosLeung1999}.
\item Texture denoising by nonlocal means \cite{BuadesCollMorel2005}.
\item Image synthesis by summation with the cartoon component $\Bu$.
\end{itemize}

The DG3PD model enforces sparsity and smoothness on the obtained texture component $\Bv$.
Sparsity means that the vast majority of coefficients in $\Bv$ are equal to zero 
(the percentage of zero coefficients depends on how much texture a specific image contains).
We make good use of this property to answer the question: which pixels of the missing shall be inpainted with texture?
First, the texture component $\Bv$ is segmented into zero (gray pixels in Figure \ref{fig:TextureMorphology}(a))  
and nonzero coefficients (black and white pixels in Figure \ref{fig:TextureMorphology}(a)).
Morphological processing (as described in \cite{ThaiGottschlich2015G3PD,ThaiHuckemannGottschlich2015} for fingerprint segmentation)
obtains texture regions (see Figure \ref{fig:TextureMorphology}(b)).
The mask of missing pixels $D$ shown in (c) is dilated (we used a circular structure element with 5 pixels radius 
in order to avoid border effects on the margin between existing and missing pixels) 
and then, it is intersected with the texture region segmentation to obtain 
the mask $I$ shown in Figure \ref{fig:TextureMorphology}(d) which are the pixels to be inpainted with texture.

The proposed texture inpainting proceeds in two phases:
First, we build a dictionary of texture patches and second,
we inpaint all pixels of mask $I$ in a specific order.
For the dictionary, we select all patches of size $s \times s$ pixels (we used $s = 15$ in our experiments)  
from texture component $\Bv$ which meet the following two criteria: 
At least $p_1$ percent of the pixels are known (and additional, the central pixel of the patch has to exist)
and at least $p_2$ percent of the coefficients are nonzero (we used $p_1 = 70\%$ and $p_2 = 60\%$ in our experiments).
The first criterion excludes patches which contain too many missing (unknown) pixels
and the second criterion excludes patches without texture from the dictionary.
Next, we iterate over all pixels of mask $I$
and consider the pixel to be inpainted as the central pixel 
of an image patch with size $s \times s$.
We count the number of known pixels inside the patch
and compute the percentage of known pixels.
Pixels are inpainted in the following order.
We start with a threshold percentage $t = 90\%$ 
which is decreased in steps of $5\%$ per iteration
and in each iteration, we inpaint all pixels of mask $I$
for which at least $t$ percent are known.
The rationale behind this ordering 
is that the more pixels are known (or already inpainted) 
in the neighborhood around a missing pixels, 
the better it can inpainted, because this additional information 
improves the chances of finding a good match in the texture dictionary. 
A third constraint ensures that the overlap of known pixels between 
a dictionary patch and the patch around the missing pixels contains at least $p_3$ percent of known pixels
(we used $p_3 = 30\%$ in our experiments).
Inpainting a pixel means that we find the best fitting patch in our texture dictionary
and set the pixel value to the central pixel of that patch.
Best fitting is defined as minimum sum of squared differences per pixel, divided by the number of pixels which overlap.

After all missing texture pixels have been inpainted 
the texture region is denoised $n$ iterations of nonlocal means.
In each iteration, we first construct a dictionary considering all patches in the texture region
with known and previously inpainted pixels.
Next, for each pixel to be denoised, we find the top $k$ fitting patches 
in the dictionary using the same distance function (sum of squared pixelwise differences)
and set the denoised pixel to the average value of the top $k$ central pixel values.
In our experiments, we used $k = 5$ and we have observed that after about $n = 10$ iterations, 
the image has reached a steady state.
See Figure \ref{fig:TextureInpaintingAndDenoising} for a visualization of the status at several intermediate steps
during the inpainting and denoising process.

Finally, the full image is reconstructed by summation of the inpainted cartoon component $\Bu$ 
with the inpainted and denoised texture component $\Bv$, see Figure~\ref{fig:DG3PD_inpainting:DiscreteInnovationModel}.

\section{Comparison of DG3PD to Further Inpainting Methods} 
\label{sec:comparisonPriorArt}

\begin{figure} 
\begin{center}    
	\includegraphics[width=1\textwidth]{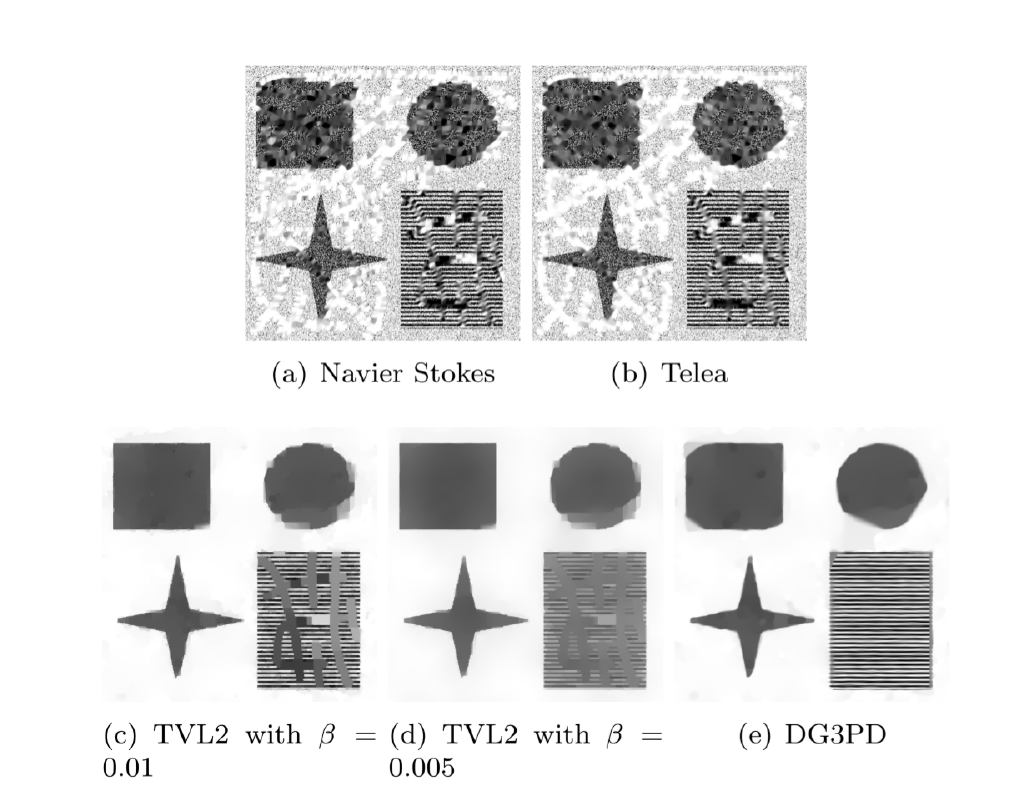}
	
  \caption{Comparison of inpainting results obtained by (a) Navier Stokes \cite{BertalmioBertozziSapiro2001}, 
	         by (b) Telea \cite{Telea2004}, by (c,d)TVL2 and by (e) DG3PD.}
  \label{fig:comparison}
\end{center}
\end{figure}

Figure \ref{fig:comparison} shows inpainting results 
for the considered challenging problem (see Figure \ref{fig:challenge} (c))
obtained by well-known inpainting methods from the literature: 
an approach based on Navier Stokes equations from fluid dynamics
which has been proposed by Bertalmio \textit{et al.} \cite{BertalmioBertozziSapiro2001},
an inpainting method suggested by Telea \cite{Telea2004}
and a well-known TVL2 inpainting approach with its synthesis image shown in Figure \ref{fig:comparison} (b) and (c).

The image shown in Figure \ref{fig:challenge}(c) has the three properties 
discussed in Section~1: 
\begin{itemize}
\item[(i)]~a large percentage of pixels in $\Bf$ are missing and shall be inpainted.
\item[(ii)]~the known pixels in $\Bf$ are corrupted by noise.
\item[(iii)]~$\Bf$ contains both cartoon and texture elements.
\end{itemize}
Furthermore, a noise-free, ideal version $\Bf_0$ depicted in Figure \ref{fig:challenge}(a) is available 
which serves as ground truth for evaluating the quality of inpainted and denoised images by different methods.
The availability of a noise-free image is an advantage for comparing and evaluating different 
inpainting and denoising methods.
In contrast, images taken by a digital camera contain a certain amount of noise
and for them, an ideal version is not available, 
see e.g. the Barbara image we used in previous comparisons \cite{ThaiGottschlich2015DG3PD}.
The choice of the ideal image (shown in Figure \ref{fig:challenge}) enables to clearly see and understand
the advantages and limitations of the compared methods (see Figure \ref{fig:comparison}).

Due to the fidelity term, the $L_2$-norm, in the TVL2 inpainting model, noise is reduced
and the homogeneous areas are well interpolated. However, the method is known to cause the ``stair case'' effect
on the homogeneous regions and texture,
while small scale objects tend to be eliminated.
If the parameter $\beta$ is chosen smaller (e.g. $\beta_5 = 0.005$), 
the resulting inpainted image is smoother, i.e. effect of ``stair case''  is reduced, 
but also more texture parts are removed.

DG3PD inpainting produces a smooth result without stair case effect.
In comparison to the other approaches, it is the only method 
which can reconstruct the texture regions in a satisfactory way.


\section{From the Bayesian Framework to the DG3PD Inpainting Model} 
\label{sec:BayesianDecomposition}

In this section, we describe the DG3PD inpainting model from a view of the Bayesian framework through maximum a posterior (MAP) estimator.
We assume that $\Bu$ and $\Bv$ are independent random variables (r.v.).
\begin{itemize}
 \item {\bfseries Prior of cartoon image $\Bu$:}
  A cartoon $\Bu = \big[ u[\Bk] \big]_{k \in \Omega}$ consists of 
  element pixel $u[\Bk]$ which is considered as an independent r.v. with
  a prior $p_{U_k}$. 
  
  Given $\Bk \in \Omega$, denote $r[\Bk] = \nabla_L^+ u[\Bk] = \Big[ r_l[\Bk] \Big]_{l=0}^{L-1}$.
  To describe the distribution of a r.v. $U_k$, we firstly define the distribution of 
  $L$-dimensional r.v. $\mathbf R_k = [R_{0,k}, \ldots, R_{L-1,k}]$.
  We assume that r.v. $\mathbf R_k$ has a multi-dimensional Laplace distribution
  which is a part of multi-dimensional power exponential distribution \cite{KotzKozubowskiPodgorski2001, KotzNadarajah2004},
  i.e. $\mathbf R_k \sim \text{PE}_L(0, \Sigma, \frac{1}{2}) = \text{Laplace}_L(0, \Sigma)$ with
  $\Sigma = \begin{pmatrix} \gamma^2 & 0 \\ 0 & \gamma^2 \end{pmatrix}$: 
  \begin{align*}
   p_{\mathbf R_k} (r[\Bk] \mid 0, \Sigma, \frac{1}{2}) &= \frac{1}{8 \pi \gamma^2} 
   \exp \Big\{ -\frac{1}{2} \Big( 
   \begin{pmatrix} r_0[\Bk] \\ \vdots \\ r_{L-1}[\Bk] \end{pmatrix}^T 
   \frac{1}{\gamma^4}
   \begin{pmatrix} \gamma^2 & 0 \\ 0 & \gamma^2 \end{pmatrix}
   \begin{pmatrix} r_0[\Bk] \\ \vdots \\ r_{L-1}[\Bk] \end{pmatrix}
   \Big)^\frac{1}{2} \Big\}
   \\
   &= \frac{1}{8\pi \gamma^2} \exp \Big\{ -\frac{1}{2\gamma}  
   \underbrace{\Big[ \sum_{l=0}^{L-1} \abs{r_l[\Bk]}^2 \Big]^\frac{1}{2}}_{:= \abs{r[\Bk]} = \abs{\nabla^+_L u[\Bk]}} 
   \Big\}.
  \end{align*}
  Thus, the distribution of a r.v. $U_k$ is a multi-dimensional Laplace distribution with operator $\nabla_L^+$:
  \begin{equation*}
   p_{U_k} (u[\Bk]) = \frac{1}{8\pi \gamma^2} \exp \Big\{ -\frac{1}{2\gamma} \abs{\nabla^+_L u[\Bk]} \Big\}.
  \end{equation*}
  The joint probability density function (p.d.f.) of a prior $\Bu$ is
  \begin{equation*}
   p_{U}(\Bu) = \prod_{\Bk \in \Omega} p_{U_k}(u[\Bk]) 
   = \frac{1}{(8\pi \gamma^2)^{\abs{\Omega}}} \exp \Big\{ -\frac{1}{2\gamma} \sum_{\Bk \in \Omega} \abs{\nabla^+_L u[\Bk]} \Big\}
   = \frac{1}{(8\pi \gamma^2)^{\abs{\Omega}}} e^{-\frac{1}{2\gamma} \norm{\nabla_L^+ \Bu}_{\ell_1}}.
  \end{equation*}  
  We choose $\gamma = \frac{1}{2}$.
  The potential function of $\Bu$ in a matrix form is
  \begin{equation*}
   \Phi_U(\Bu) = - \log p_U(\Bu) = \norm{\nabla_L^+ \Bu}_{\ell_1} + \abs{\Omega} \log (2\pi).
  \end{equation*}
  Note that the original $\Bu$ is not a Laplace distribution, but a transform of $\Bu$ under an operator $\nabla^+_L$
  has an independent multi-dimensional Laplace distribution.

 \item {\bfseries Prior of texture image $\Bv$:} 
  Under a transform operator $\mathcal T$, the texture image $\Bv$ is decomposed in different orientations.
  Operator $\mathcal T$ is suitably chosen to capture the texture components in the original image $\Bf$.
  As definition of discrete multi-dimensional G-norm \cite{ThaiGottschlich2015DG3PD}, 
  the definition of the transform $\mathcal T$ 
  (in a direction $s$, $s = 0, \ldots, S-1$) is defined by its inversion
  (note that $\mathcal T \mathcal T^{-1} = \text{Id}$):
  \begin{equation*}
   \mathcal T \{ \Bv \} = \big[ \underbrace{\mathcal T_s \{\Bv\}}_{:= \Bg_s} \big]_{s=0}^{S-1} = \Bg
   ~\Leftrightarrow~
   \Bv = \mathcal T^{-1} \Big\{ \big[ \Bg_s \big]_{s=0}^{S-1} \Big\}
   = \text{div}^+_S \Bg \,.
  \end{equation*}
  We assume that a texture $\Bv$ is sparse in a transform domain $\mathcal T$ and sparse in the spatial domain
  (nonzero coefficients of $\Bv$ are only due to texture).
  To satisfy these conditions, we assume that a r.v. $V_k$
  has a mixture of a Laplace distribution in spatial domain with a p.d.f. $p_{V_k,2}(v[\Bk])$ 
  (w.r.t. sparse in spatial domain) and multi-dimensional Laplace distribution (in an operator $\mathcal T$) 
  with a p.d.f. $p_{V_k,1}(v[\Bk])$ (w.r.t. sparse in transform domain $\mathcal T$).
  Thus, we have a p.d.f. of r.v. $V_k$ as follows
  \begin{equation} \label{eq:pdf:v}
   p_{V_k} (v[\Bk]) = p_{V_{1k}}(v[\Bk]) \cdot p_{V_{2k}}(v[\Bk]).
  \end{equation}
  Since $V_{2k} \sim \text{Laplace}(0, \gamma_1^2)$, we have
  \begin{equation} \label{eq:pdf:v2}
   p_{V_{2k}}(v[\Bk]) = \frac{1}{4\gamma_2} \exp \Big\{ -\frac{1}{2 \gamma_2} \abs{v[\Bk]} \Big\}.
  \end{equation}
  To define the distribution of a r.v. $V_{1k}$, we define the distribution of $S$-dimension r.v.
  $\mathbf G_k = [G_{0,k}, \ldots, G_{L-1,k}]$.
  Similar to $\mathbf U_k$, we assume that r.v. $\mathbf G_k$ has a multi-dimensional Laplace distribution, 
  i.e. $\mathbf G_k \sim \text{Laplace}_S(0, \Sigma)$ with $\Sigma = \begin{pmatrix} \gamma_1^2 & 0 \\ 0 & \gamma_1^2 \end{pmatrix}$.
  It is easy to see that (with $g[\Bk] = \cT\{\Bv\}[\Bk] \,, \Bk \in \Omega$)
  \begin{equation*}
   p_{\mathbf G_k} (g[\Bk]) = \frac{1}{8\pi\gamma_1^2} \exp\Big\{ -\frac{1}{2\gamma_1} \abs{g[\Bk]} \Big\}
  \end{equation*}
  or
  \begin{equation} \label{eq:pdf:v1}
   p_{V_{1k}} (v[\Bk]) = \frac{1}{8\pi\gamma_1^2} \exp\Big\{ -\frac{1}{2\gamma_1} \abs{\cT\{\Bv\}[\Bk]} \Big\}.
  \end{equation}  
  Put (\ref{eq:pdf:v2}) and (\ref{eq:pdf:v1}) to (\ref{eq:pdf:v}), we have the p.d.f. of r.v. $V_k$ as
  \begin{equation*}
   p_{V_k} (v[\Bk]) = \frac{1}{32\pi\gamma_1^2 \gamma_2} \exp\Big\{ -\frac{1}{2\gamma_1} \abs{\cT\{\Bv\}[\Bk]} - \frac{1}{2 \gamma_2} \abs{v[\Bk]} \Big\}.
  \end{equation*}
  The joint p.d.f. of a texture image $\Bv = \big[ v[\Bk] \big]_{\Bk \in \Omega}$ is
  \begin{equation*}
   p_V(\Bv) = \prod_{\Bk \in \Omega} p_{V_k}(v[\Bk]) = 
   \frac{1}{ (32\pi\gamma_1^2 \gamma_2)^{\abs{\Omega}} }
   \exp \Big\{ -\frac{1}{2 \gamma_1} \norm{\cT\{\Bv\}}_{\ell_1} - \frac{1}{2 \gamma_2} \norm{\Bv}_{\ell_1} \Big\}.
  \end{equation*}
  we choose $\mu_2 = \frac{1}{2 \gamma_2}$ and $\mu_1 = \frac{1}{2 \gamma_1}$.
  The potential function of $\Bv$ with an anisotropic version 
  $ \big( \norm{\mathcal T\{\Bv\}}_{\ell_1} = \sum_{s=0}^{S-1} \norm{ \mathcal T_s\{\Bv\} }_{\ell_1}  = \sum_{s=0}^{S-1} \norm{ \mathcal \Bg_s }_{\ell_1} \big)$ 
  in a matrix form is defined as
  \begin{equation*}
   \Phi_V(\Bv) = \mu_1 \sum_{s=0}^{S-1} \norm{\Bg_s}_{\ell_1} + \mu_2 \norm{\Bv}_{\ell_1} + \abs{\Omega} \log \frac{4 \pi}{\mu_1^2 \mu_2} \,.
  \end{equation*}

 \item {\bfseries Likelihood (or the joint p.d.f. of $p_{F \mid U, V} (\Bf \mid \Bu, \Bv)$):} 
  If we assume that the residual $\Beps$ in an image $\Bf$ has a power exponential distribution 
  \cite{KotzKozubowskiPodgorski2001, KotzNadarajah2004}, i.e. $E_k \sim \text{PE}(\mu, \sigma^2, \xi)$, 
  e.g. Gaussian $(\xi = 1)$ or Laplacian $(\xi = \frac{1}{2})$, its density function at $\Bk \in \Omega$ is   
  \begin{equation*}
   p_{E_k} (\epsilon[\Bk] \mid \mu) = \frac{1}{\Gamma(1+\frac{1}{2\xi}) 2^{1+\frac{1}{2\xi}} \sigma} 
   \exp \Big\{ -\frac{1}{2} \frac{\abs{\epsilon[\Bk]-\mu}^{2 \xi}}{\sigma^{2\xi}} \Big\}.
  \end{equation*}
  Due to the inpainting problem with a missing region $D$, 
  the likelihood is defined on a known domain $\Omega \backslash D$ as
  \begin{align*}
   p_{F \mid U, V} (\Bf \mid \Bu, \Bv) &= \prod_{\Bk \in \Omega} p_{F_k \mid U_k, V_k} (f[\Bk] \mid u[\Bk], v[\Bk])
   \\
   &= \frac{1}{\Big( \Gamma(1+\frac{1}{2\xi}) 2^{1+\frac{1}{2\xi}} \sigma \Big)^{\abs{\Omega}}}
   \exp \Big\{ -\frac{1}{2 \sigma^{2\xi}} \underbrace{ \sum_{\Bk \in \Omega} \abs{\chi_D^c[\Bk] ( f[\Bk]-u[\Bk]-v[\Bk] )}^{2 \xi} }_{ := \norm{\chi_D^c \cdot^\times (\Bf - \Bu - \Bv)}^{2\xi}_{2\xi} } \Big\} \,.
  \end{align*}
  
 \item {\bfseries A posterior:}
 Since $\Bu$ and $\Bv$ are independent to each other, a posterior is written as
  \begin{equation*}
   p_{U,V \mid F} (\Bu, \Bv \mid \Bf) = \frac{ p_{F \mid U,V}(\Bf \mid \Bu, \Bv) p_U(\Bu) p_V(\Bv) }{p_F(\Bf)} 
   \propto p_{F \mid U,V}(\Bf \mid \Bu, \Bv) p_U(\Bu) p_V(\Bv) \,.
  \end{equation*}  
\end{itemize}
Let $\Beps = \Bf - \Bu - \Bv$, the MAP estimator, i.e. $\max_{(\Bu, \Bv) \in X^2} p_{U,V \mid F} (\Bu, \Bv \mid \Bf)$, is defined as
\begin{align*}
 &\min_{(\Bu, \Bv) \in X^2} \Big\{ \underbrace{- \log p_U(\Bu)}_{= \Phi_U(\Bu)} ~ \underbrace{- \log p_V(\Bv)}_{= \Phi_V(\Bv)} 
  -\log p_{U,V \mid F} (\Bu, \Bv \mid \Bf) \Big\}
 \\
 = &\min_{(\Bu, \Bv) \in X^2} \Big\{ \norm{\nabla_L^+ \Bu}_{\ell_1} + \mu_1 \sum_{s=0}^{S-1} \norm{\Bg_s}_{\ell_1} + \mu_2 \norm{\Bv}_{\ell_1}
    + \frac{1}{2 \sigma^{2\xi}} \norm{\chi_D^c \cdot^\times \Beps}^{2\xi}_{2\xi} \,,
 \\ & \qquad \qquad \text{s.t.}~~
    \Bv = \sum_{s=0}^{S-1} \big[ \cos(\frac{\pi s}{S}) \Bg_s \BDnT + \sin(\frac{\pi s}{S}) \BDm \Bg_s \big] 
    \,,  \Bf = \Bu + \Bv + \Beps \Big\} .
\end{align*}
However, in practice, we do not know which types of noise are observed in a signal. 
Instead of the $\ell_{2 \xi}$ norm of the residual $\Beps$ on a known domain $\Omega \backslash D$ to characterize the properties of noise,
we control the smoothness of the solution by the maximum of curvelet coefficient of $\Beps$ 
on a domain $\Omega \backslash D$ by a constant $\nu$, 
i.e. $\norm{\mathcal C \{\chi^c_D \cdot^\times \Beps \}}_{\ell_\infty} \leq \nu$.

In \cite{HaltmeierMunk2014}, if noise in image $\Bf$ is Gaussian, as in extreme value theory, 
the r.v. $\norm{\mathcal C \{\chi^c_D \cdot^\times \Beps \}}_{\ell_\infty}$ has a Gumbel distribution.
Thus, there is a condition to choose $\nu$.
However, in practice, if noise cannot be identified, $\nu$ is chosen by $\alpha$-quantile.
Note that the condition $\norm{\mathcal C \{\chi^c_D \cdot^\times \Beps \}}_{\ell_\infty} \leq \nu$ is similar to
the Dantzig selector \cite{CandesTao2007}. 

\begin{figure}
\begin{center}    
  
	\includegraphics[width=0.95\textwidth]{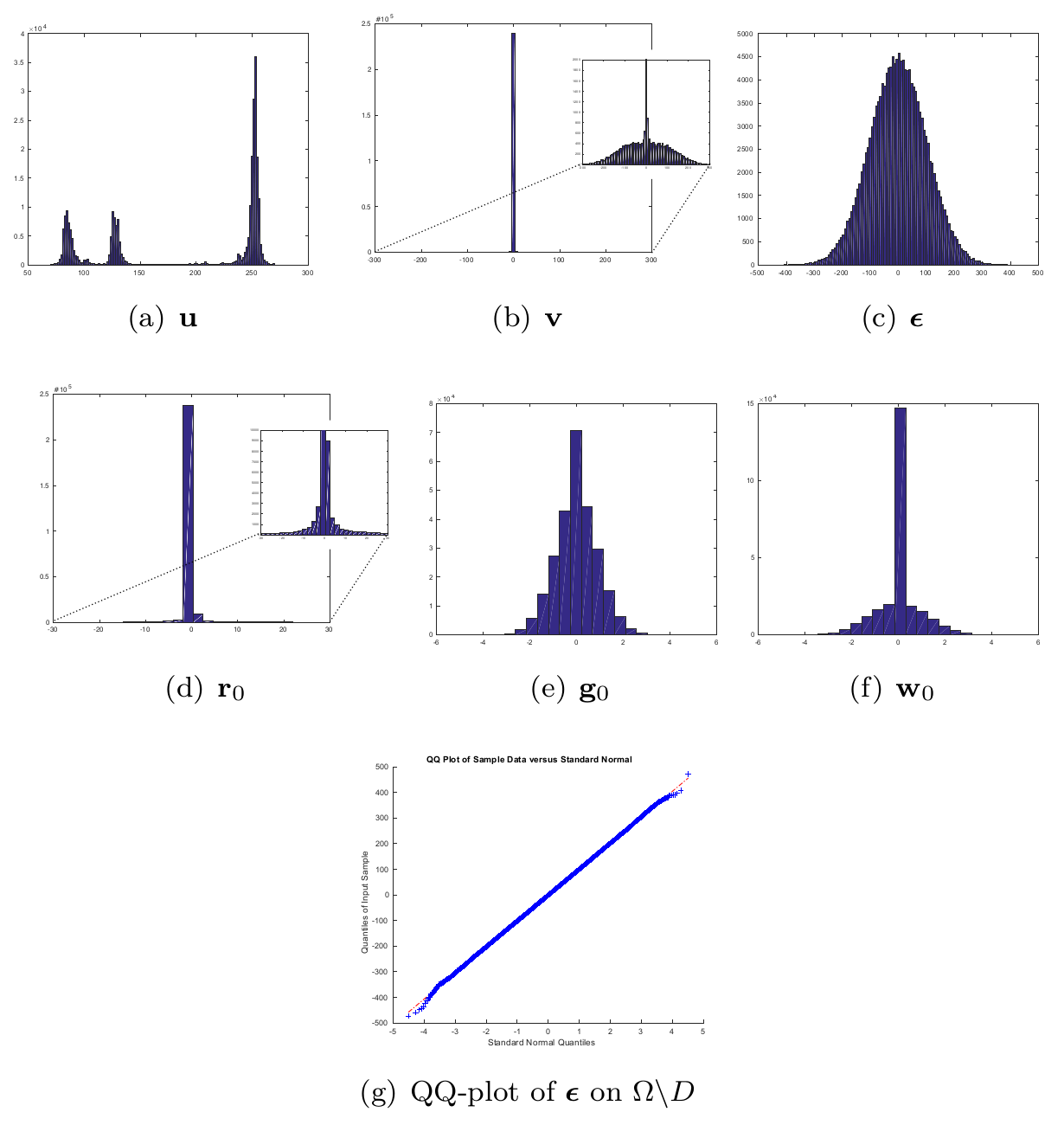}
	
  \caption{      
           The figure depicts the empirical density functions of the solution in 
           (\ref{eq:DG3PD_inpainting_ALM_numerical}) by ADMM:        
           (a), (b), (c), (d), (e), (f) and (g) are the empirical density functions of 
           $\Bu \,, \Bv \,, \Beps \,, \mathbf{r_0} \,, \mathbf{g_0} \,, \mathbf{w_0}$
           and the QQ-plot of the residual $\Beps$, respectively.
           We assume that the prior of $\nabla_L^+ \Bu$ has a multi-Laplacian distribution,
           i.e. a changed variable $\Br = \big[ \Br_l \big]_{l=0}^{L-1} = \nabla_L^+ \Bu$ 
           has a multi-Laplace distribution,
           or $\Br_l$ has an univariate Laplace distribution.
           This effect causes the sparseness of $\Br_l$ due to the shrinkage operator, see (d).          
           The sparsity for $\nabla_L^+ \Bu$ causes the smoothness of the objects in the cartoon $\Bu$, see (a).           
           The smooth and sparse texture $\Bv$ (due to the expansion of a whole range in intensity value of its non-zero coefficients
           caused by $\norm{\Bv}_{\ell_1}$ in (\ref{eq:DG3PD_inpainting})) is illustrated in (b).
           Because of an additive white Gaussian noise in a simulation,
           the distribution of $\Beps$ in $\Omega \backslash D$ 
           is approximately normal, see plot (c) and its QQ-plot (g).
           The reasons of the differences near the boundary in (g) may cause by: (1) a simulation of a Gaussian noise,
           i.e. the tail of a Gaussian noise does not go to infinity in a simulation. 
           (2) the remaining texture in $\Beps$ (due to a selection of $\nu$). 
           Changing variable $\Bw = \Bg$ causes a non-sparsity in $\Bg$ and a sparsity in $\Bw$ (which is shrinked from $\Bg$), c.f. (e) and (f).
           }
           
\label{fig:DG3PD:densityFunc}
\end{center}
\end{figure}

Figure \ref{fig:DG3PD:densityFunc} illustrates the empirical density functions of the solution of the minimization 
(\ref{eq:DG3PD_inpainting_ALM_numerical}). The statistical properties of the solution can be characterized
in the Bayesian framework with the priors, likelihood and posterior as mentioned above.

\begin{figure} 
\begin{center}    
  \includegraphics[width=1.1\textwidth]{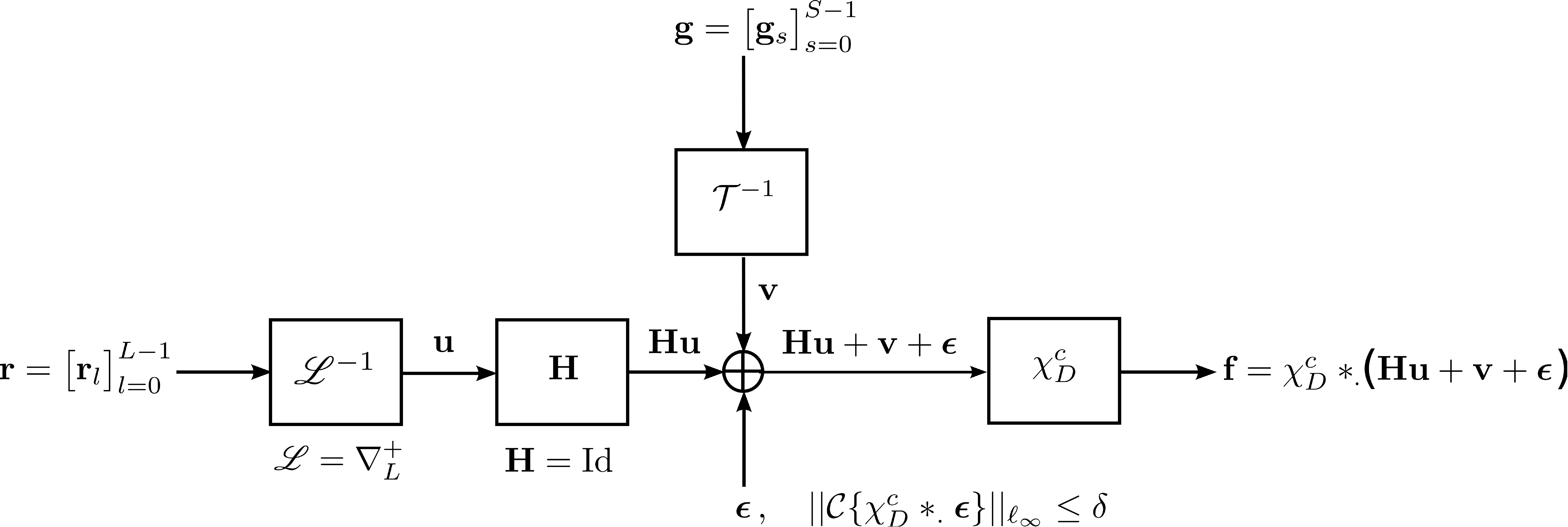}    
  \caption{Overview over the discrete innovation model for the DG3PD based image inpainting and denoising.
           Note that the $\mathscr L^{-1}$ does not exist in general, but the reconstructed image is obtained 
           by the function space of the inverse problem.
           \label{fig:DG3PD_inpainting:DiscreteInnovationModel} }
\end{center}
\end{figure}

Note that the DG3PD inpainting model (\ref{eq:DG3PD_inpainting}) can be described as an inverse problem
in image restoration (e.g. image inpainting and denoising), 
see Figure \ref{fig:DG3PD_inpainting:DiscreteInnovationModel} for the discrete innovation model.

\section{Relation Between Variational Analysis and Pyramid Decomposition} \label{sec:VariationalFourierConnection}

In this section, we discuss connections and similarities between the proposed DG3PD inpainting model
and existing work in the area of 
pyramid representations, multiresolution analysis, and scale-space representation.
Historically, Gaussian pyramids and Laplacian pyramids have been developed for
applications like texture synthesis, image compression and denoising.
Early works in this area include a paper by Burt and Adelson in 1983 \cite{BurtAdelson1983} applying Laplacian pyramids for image compression.
Pyramidal decomposition schemes \cite{BurtAdelson1983,SimoncelliFreeman1995,UnserChenouardVandeville2011} can be grouped into two categories.
Their main property is either lowpass or bandpass filtering.
Very related are transforms from multiresolution analysis, 
including wavelet, 
curvelet \cite{CandesDonoho2004, CandesDemanetDonohoYing2006}, 
contourlet \cite{DoVetterli2005}, and steerable wavelet \cite{UnserVandeville2010}.
A difference of the proposed DG3PD inpainting model is that its 
basis elements for obtaining the decomposition
are constructed 
by discrete differential operators
and the decomposition is solved in a non-linear way
which adapts to each image (enabled by update iterations, corresponding to the loop in Figure \ref{fig:VariationPyramid}).
The discussion of commonalities and differences is made more precise and detailed in the following rest of the section.

Let denote $\stackrel{\cF}{\longleftrightarrow}$ a discrete Fourier transform pair.
Given $z_1 = e^{j \omega_1}$ and $z_2 = e^{j \omega_2}$ and the Dirac delta function $\delta(\cdot)$, 
the impulse response of the discrete directional derivative operators ($l = 0 \,, \ldots \,, L-1$) and their spectra are
\begin{itemize}
 \item the forward operator: 
  \begin{equation*}
   \partial_l^+ \delta[\Bk] = \Big[ \cos(\frac{\pi l}{L}) \partial_x^+  +  \sin(\frac{\pi l}{L}) \partial_y^+ \Big] \delta[\Bk]
   ~\stackrel{\cF}{\longleftrightarrow}~ \cos(\frac{\pi l}{L}) (z_2 - 1) + \sin(\frac{\pi l}{L}) (z_1 - 1) \,,
  \end{equation*}
  
 \item the backward operator: 
  \begin{equation*}
   \partial_l^- \delta[\Bk] = \Big[ \cos(\frac{\pi l}{L}) \partial_x^-  +  \sin(\frac{\pi l}{L}) \partial_y^- \Big] \delta[\Bk]
   ~\stackrel{\cF}{\longleftrightarrow}~ -\cos(\frac{\pi l}{L}) (z_2^{-1} - 1) - \sin(\frac{\pi l}{L}) (z_1^{-1} - 1) \,.
  \end{equation*}
\end{itemize}

\subsection{The ``$\Bu$-problem''}

In order to describe the variational analysis for the $\Bu$-problem
from the view of filtering in the Fourier domain for a pyramid decomposition scheme, 
we define the discrete inverse Fourier transform of $\mathcal X(\Bz)$ and $\mathcal Y(\Bz)$ in 
the ``$\Bu$-problem'' in \cite{ThaiGottschlich2015DG3PD} as
\begin{align*}
 &\mathcal X(\Bz) ~\stackrel{\mathcal F^{-1}}{\longleftrightarrow}~
 X[\Bk] = \Big[ \beta_4 - \beta_1 \sum_{l=0}^{L-1} \partial_l^- \partial_l^+ \Big] \delta [\Bk],
 \\
 &\mathcal Y(\Bz) ~\stackrel{\mathcal F^{-1}}{\longleftrightarrow}~
 Y[\Bk] = \beta_4 \big( f[\Bk] - v[\Bk] - \epsilon[\Bk] + \frac{\lambda_4[\Bk]}{\beta_4} \Big)
 - \beta_1 \sum_{l=0}^{L-1} \partial_l^- \Big\{ 
 \underbrace{ \Shrink \Big( \partial_l^+ u[\Bk] - \frac{\lambda_{\boldsymbol{1} l}[\Bk]}{\beta_1} \,, \frac{1}{\beta_1} \Big) }_{:=~ r_l[\Bk]}
 + \frac{\lambda_{1l}[\Bk]}{\beta_1} \Big\} \,.
\end{align*}
Note that the function $\mathcal X(\Bz)$ satisfies
$$ 
0 < \beta_4 \leq \mathcal X (\Bz) < +\infty \,, \forall \Bome \in [-\pi \,, \pi]^2
$$
and it is similar to the autocorrelation function in \cite{Unser2000}
which follows the condition of the Riesz basis.
We rewrite cartoon $\Bu$ in the Fourier and spatial domain in the scheme of filter design as
\begin{align} \label{eq:projection:u}
 U(\Bz) &= \mathcal X^{-1}(\Bz) \cdot \mathcal Y(\Bz)
 = \Phi(\Bz) \Big( F(\Bz) - V(\Bz) - E(\Bz) + \frac{\Lambda_4(\Bz)}{\beta_4} \Big)
    ~+~ \sum_{l=0}^{L-1} \tilde \Psi(\Bz) \Big[ R_l(\Bz) + \frac{\Lambda_{1l}(\Bz)}{\beta_1} \Big]   \notag
 \\
 \stackrel{\mathcal F^{-1}}{\longleftrightarrow}
 u[\Bk] &= \underbrace{ \Big( \phi*(f-v-\epsilon + \frac{\lambda_4}{\beta_4}) \Big)[\Bk] }_{:= u_\text{update} [\Bk]}
 ~+~ \underbrace{ \sum_{l=0}^{L-1} \Big( \tilde \psi_l \ast \Big[ \Shrink \Big( \psi_l \ast u - \frac{\lambda_{1 l}}{\beta_1} \,, \frac{1}{\beta_1} \Big)
 + \frac{\lambda_{1l}}{\beta_1} \Big] \Big)[\Bk]   
 }_{:=~ \text{FST}_{\psi, \tilde \psi, c_1 }( u \,, \frac{\lambda_{1 l}}{\beta_1} \,, \frac{1}{\beta_1})  ~:= u_\text{smooth} [\Bk] 
 ~~ \text{(frame soft thresholding)}}
\end{align}
Given $\beta_1 = c_1 \beta_4$, the spectra and impulse responses of the scaling and frame functions in (\ref{eq:projection:u}) 
$(l = 0 \,, \ldots \,, L-1)$ are
\begin{align*}
 \Phi(\Bz) &= \beta_4 \mathcal X^{-1}(\Bz) 
 ~\stackrel{\mathcal F}{\longleftrightarrow}~ 
 \phi[\Bk] = \Big[ 1 - c_1 \sum_{l=0}^{L-1} \partial_l^- \partial_l^+ \Big]^{-1} \delta [\Bk] \,,
 \\
 \tilde \Psi_l(\Bz) &= \beta_1 \mathcal X^{-1}(\Bz) \Big[ (z_1^{-1}-1) \sin \frac{\pi l}{L} ~+~ (z_2^{-1}-1) \cos \frac{\pi l}{L} \Big]
 ~\stackrel{\mathcal F}{\longleftrightarrow}~ 
 \tilde \psi_l[\Bk] = - c_1 \Big[ 1 - c_1 \sum_{l=0}^{L-1} \partial_l^- \partial_l^+ \Big]^{-1} \partial_l^- \delta [\Bk] \,,
 \\
 \Psi_l (\Bz) &= (z_1 - 1) \sin \frac{\pi l}{L}  + (z_2 - 1) \cos \frac{\pi l}{L}
 ~\stackrel{\mathcal F}{\longleftrightarrow}~ 
 \psi_l [\Bk] = \partial_l^+ \delta [\Bk].
\end{align*}
The equation (\ref{eq:projection:u}) is somewhat similar to the projection operator 
in the pyramid decomposition scheme (see curvelet \cite{CandesDonoho2004, CandesDemanetDonohoYing2006}, 
contourlet \cite{DoVetterli2005}, steerable wavelet \cite{UnserVandeville2010}, etc.), with:
\begin{itemize}
 \item scaling function $\varphi(\cdot)$ and its dual $\tilde \varphi(\cdot)$ with the interpolant $\phi = \varphi \ast \tilde \varphi $,
 \item frame function $\psi_l(\cdot)$ and its dual $\tilde \psi_l(\cdot)$,
 \item proximity operator for frame elements, i.e. $\Shrink(\cdot \,, \cdot)$.
\end{itemize}
Note that the frame soft thresholding $\text{FST}(\cdot, \cdot, \cdot)$ in (\ref{eq:projection:u})
consists of frame element $\psi_l$, its dual $\tilde \psi_l$
and shrinkage operator $\Shrink(\cdot \,, \cdot)$ together with the updated Lagrange multiplier.
Although $\text{FST}(\cdot, \cdot, \cdot)$ is similar to wavelet/curvelet shrinkage operator, 
this operator is obtained from the calculus of variation.

\begin{figure}[ht] 
\begin{center}
 \includegraphics[width=1\textwidth]{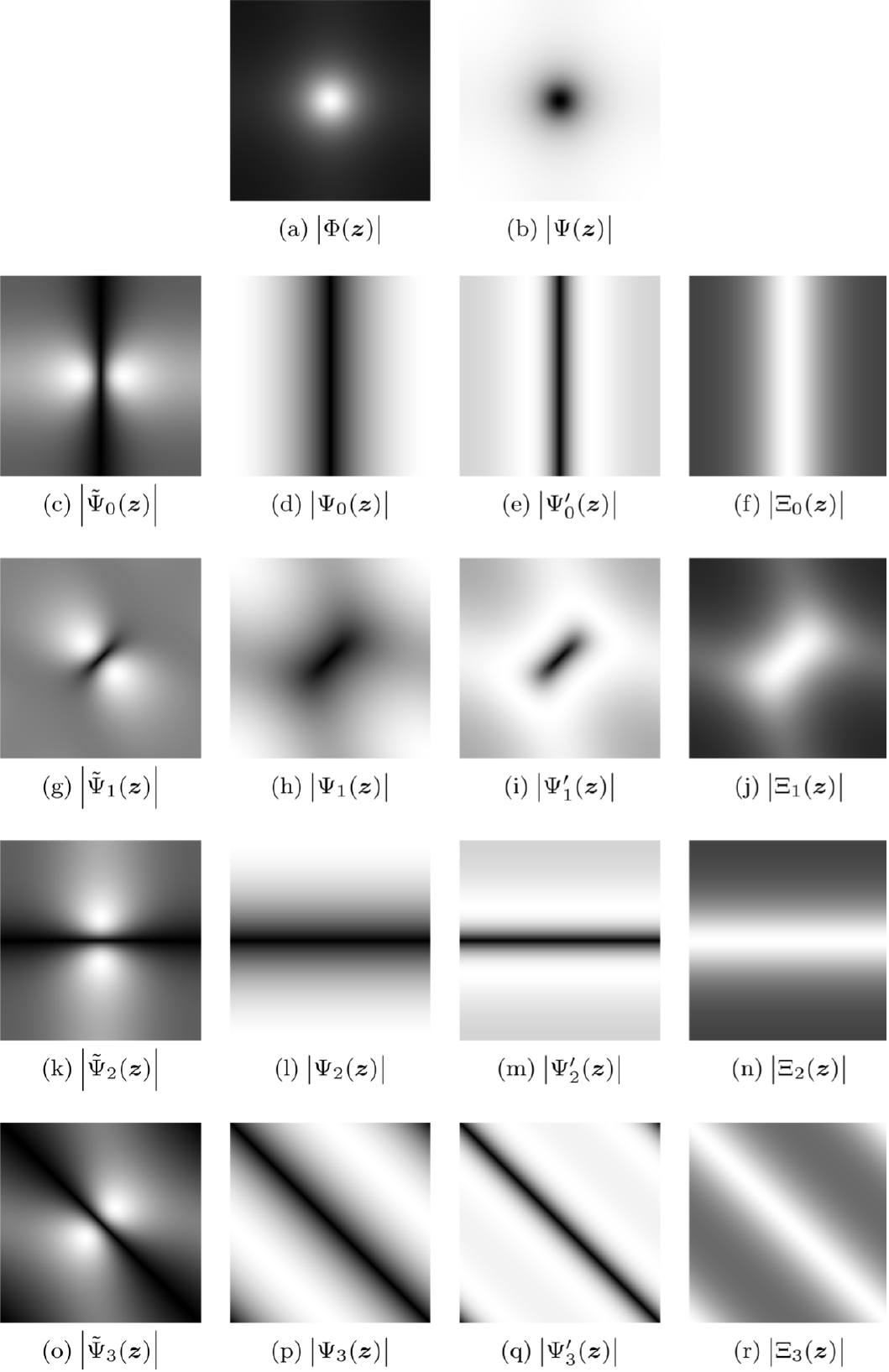}
 
 \caption{The spectra of frame functions in the ``$\Bu$-problem'' and the ``$\Bg$-problem''
          with the parameters:
          $L = S = 4 \,, \beta_1 = \beta_4 = 0.04 \,, \beta_2 = \beta_3 = 0.3$
          and $\displaystyle \Psi(\Bz) = \sum_{l=0}^{L-1} \tilde \Psi_l(\Bz) \Psi_l(\Bz)$.
          The unity properties for these frame functions are $\displaystyle \Phi(\Bz) + \sum_{l=0}^{L-1} \tilde \Psi_l (\Bz) \Psi_l (\Bz) = 1$
          and $\displaystyle \Xi_a(\Bz) + \Psi_a(\Bz) \Psi'_a(\Bz) = 1 \,,~ a = 0, \ldots , S-1$.
         }
 
 \label{fig:FourierU_L4}
\end{center}
\end{figure}

It is easy to obtain the unity property of frame elements 
$(\phi \,, \psi_l \,, \tilde \psi_l)$ in the Fourier domain as
\begin{align*}
 &\Phi(\Bz) + \sum_{l=0}^{L-1} \tilde \Psi_l(\Bz) \Psi_l(\Bz) = 1 \,,
 \\
 & \Phi(e^{j \mathbf 0}) = 1 ~~\text{and}~~ \tilde \Psi_l(e^{j \mathbf 0}) = \Psi_l(e^{j \mathbf 0}) = 0 \,,
\end{align*}
which satisfies the condition of the perfect reconstruction in the scheme of wavelet pyramid,
see Figure \ref{fig:FourierU_L4} for illustration with $L = 4$ directions.
When the number of directions $L$ increases, the bandwidth of the scaling function $\Phi(z)$ in the spectral domain reduces,
see Figure \ref{fig:FourierU_L8}.
Due to the effect of this lowpass filter, the cartoon $\Bu$ is smoother. 
These frame elements $(\phi, \psi_l, \tilde \psi_l)$ which consist of partial differential operators 
can be considered as a kind of a wavelet-like operator \cite{KhalidovUnser2006}.
However, the procedure to obtain these elements is from the calculus of variation.
\setcounter{subfigure}{0}

\subsection{The ``$\big[ \Bg_s \big]_{s=0}^{S-1}$-problem'':}

In analogy to the $\Bu$-problem in the variational framework,
we define the frame operators to extract the directional features of the texture $\Bv$ in pyramid decomposition scheme.
The discrete inverse Fourier transforms of $\mathcal A(\Bz)$ and $\mathcal B(\Bz)$ in 
the solution of the ``$\Bv$-problem'' \cite{ThaiGottschlich2015DG3PD} are defined as
\begin{align*}
 \mathcal A_a(\Bz) &~\stackrel{\mathcal F^{-1}}{\longleftrightarrow}~ 
 A_a[\Bk] ~=~ \Big[ \beta_2 - \beta_3 \partial_a^- \partial_a^+ \Big] \delta[\Bk] \,,
 \\
 \mathcal B_a(\Bz) &~\stackrel{\mathcal F^{-1}}{\longleftrightarrow}~
 B_a[\Bk] ~=~ \beta_2 \Big[ \Shrink \Big( g_a[\Bk] - \frac{\lambda_{2a}[\Bk]}{\beta_2} \,, \frac{\mu_1}{\beta_2} \Big)
 + \frac{\lambda_{2a}[\Bk]}{\beta_2} \Big]
 - \beta_3 \partial_a^- \Big\{ \underbrace{ v[\Bk] - \sum_{s=[0, S-1] \backslash \{a\}} \partial_s^+g_s[\Bk] }_{\displaystyle :=~ \partial_a^+ g_a[\Bk]}
 + \frac{\lambda_3 [\Bk]}{\beta_3} \Big\} 
\end{align*}
Similar to $\mathcal X(\Bz)$, the ``autocorrelation'' function $\mathcal A_a(\Bz)$ is bounded as 
in the condition of the Riesz basis \cite{Unser2000}, i.e.
$$ 
0 < \beta_2 \leq \mathcal A_a (\Bz) < +\infty \,, \forall \Bome \in [-\pi \,, \pi]^2 \,.
$$
Texture $\Bv$ is rewritten in the Fourier and spatial domain with $a = 0 \,, \ldots \,, S-1$, as
\begin{align} \label{eq:projection:g}
 G_a(\Bz) &~=~ \mathcal A_a^{-1}(\Bz) \cdot \mathcal B_a(\Bz) ~=~
 \Xi_a(\Bz) \Big[ W_a(\Bz)  + \frac{\Lambda_{2a}(\Bz) }{\beta_2} \Big]
 ~+~ \Psi'_a(\Bz)
 \Big[ \Psi_a (\Bz) G_a(\Bz) + \frac{\Lambda_3(\Bz)}{\beta_3} \Big]     \notag
 \\
 \stackrel{\mathcal F^{-1}}{\longleftrightarrow}~ 
 g_a[\Bk] &~=~ \bigg( \xi_a \ast \Big[ 
 \underbrace{ \Shrink \Big( g_a - \frac{\lambda_{2a}}{\beta_2} \,, \frac{\mu_1}{\beta_2} \Big)
 + \frac{\lambda_{2a}}{\beta_2} }_{\displaystyle :=~ \text{ST} \big(g_a \,, \frac{\lambda_{2a}}{\beta_2} \,, \frac{\mu_1}{\beta_2} \big)} \Big] \bigg) [\Bk]
 ~+~ \bigg( \psi'_a \ast  \Big[ \psi_a \ast g_a  + \frac{\lambda_3}{\beta_3} \Big] \bigg)[\Bk] \,.
\end{align}
Given $\beta_2 = c_2 \beta_3$, the spectra and impulse responses of the frame functions in (\ref{eq:projection:g}) are
\begin{align*}
 \Xi_a(\Bz) &= \beta_2 \mathcal A^{-1}(\Bz) 
 ~\stackrel{\mathcal F^{-1}}{\longleftrightarrow}~ 
 \xi_a[\Bk] = c_2 \big( c_2 - \partial_a^- \partial_a^+ \big)^{-1} \delta[\Bk] \,,
 \\
 \Psi'_a(\Bz) &= \beta_3 \mathcal A^{-1}(\Bz) \Big[ (z_1^{-1} - 1) \sin \frac{\pi a}{S} ~+~ (z_2^{-1} - 1) \cos \frac{\pi a}{S} \Big]
 ~\stackrel{\mathcal F^{-1}}{\longleftrightarrow}~ 
 \psi'_a[\Bk] = - \big( c_2 - \partial_a^- \partial_a^+ \big)^{-1} \partial_a^- \delta[\Bk]
 \\
 \Psi_a(\Bz) &= (z_2 - 1) \cos \frac{\pi a}{S} ~+~ (z_1 - 1) \sin \frac{\pi a}{S}
 ~\stackrel{\mathcal F^{-1}}{\longleftrightarrow}~ 
 \psi_a[\Bk] = \partial_a^+ \delta [\Bk] \,.
\end{align*}

Similar to the $\Bu$-problem, it is easy to obtain the unity property of frame elements 
$(\xi \,, \psi_a \,, \psi'_a)$ with $a \in [0 \,, S-1]$ in Fourier domain: 
\begin{align*}
 &\Xi_a(\Bz) + \Psi_a(\Bz) \Psi'_a(\Bz) = 1 \,,
 \\
 & \Xi_a(e^{j \mathbf 0}) = 1 ~~\text{and}~~ \Psi_a(e^{j \mathbf 0}) = \Psi'_a(e^{j \mathbf 0}) = 0 
\end{align*}
which satisfies the condition of the perfect reconstruction, 
see Figure \ref{fig:FourierU_L4} with $L = 4$ directions
and Figures \ref{fig:FourierU_L8} and \ref{fig:FourierGa_L8} with $L = 8$ directions.

\subsection{The ``$\Bv$-problem'':}
The solution of the $\Bv$-problem is rewritten as
\begin{equation*}
 \Bv^* = \Shrink \Big( \mathbf{t_v} \,,~ c_{\mu_2} \cdot \max_{\Bk \in \Omega} \big( \abs{t_\Bv[\Bk]} \big) \Big) 
\end{equation*}
with
\begin{equation*}
 \mathbf{t_v} = \theta \Big[ \underbrace{\sum_{s=0}^{S-1} \psi_s \ast g_s - \frac{\boldsymbol{\lambda_3}}{\beta_3}}_{\mathbf{t}_{\mathbf v \text{smooth}}} \Big]
 + (1-\theta) \Big[ \underbrace{ \Bv + \frac{\boldsymbol{\lambda_4}}{\beta_4} }_{\mathbf{t}_{\mathbf v \text{update}}} \Big] \,.
\end{equation*}
Note that there is a shrinkage operator with two terms in texture $\Bv$, 
namely a smoothing term and a updated term, balanced by a parameter $\theta$.

\subsection{The ``$\Be$-problem'':}
\begin{figure}[ht] 
\begin{center}
 \includegraphics[width=1\textwidth]{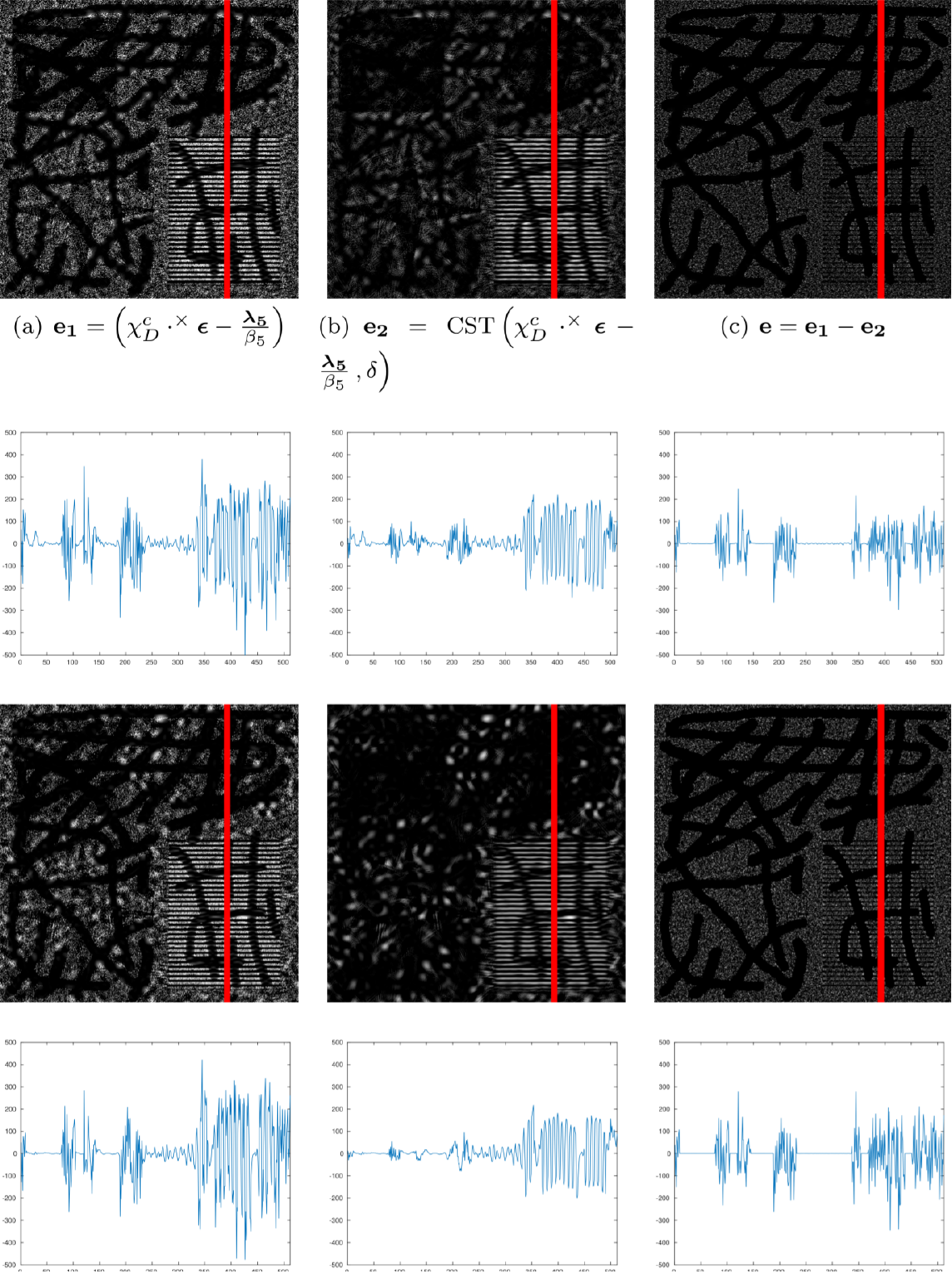}
 
 \caption{The figure illustrates the effect of a noise measurement 
          $\norm{\cC\{\chi^c_D \cdot^\times \Beps\}}_{\ell_\infty}$ on $\Omega \backslash D$
          in (\ref{eq:DG3PD_inpainting}) by introducing a new variable $\Be = \chi^c_D \cdot^\times \Beps$.
          The first and second rows are the images $\mathbf{e_1}, \mathbf{e_2}$ and $\Be$ 
          and their 1D signals along the red lines at iteration 20, respectively.
          Similarly, the third and fourth rows are at iteration 100.
          We observe that cartoon and texture still remain in $\mathbf{e_1}$ and
          its curvelet smoothing term $\mathbf{e_2}$ at iteration 20. However, these geometrical signals are reduced
          in their difference $\Be$ due to a subtraction operator.
          At iteration 100, there is almost no cartoon in $\mathbf{e_1}, \mathbf{e_2}$ and $\Be$.
          Note that there are some information in the missing domain $D$ due to the updated 
          $\boldsymbol{\lambda_5}$.
          }
 \label{fig:L_infty_Residual}
\end{center}
\end{figure}

The noise term $\norm{\cC\{\chi_D^c \cdot^\times \Beps\}}_{\ell_\infty}$ in Eq. (\ref{eq:DG3PD_inpainting})
is suitable to capture high oscillating patterns, especially small-scale types of noise,
due to the advantage of the multi-orientation and multi-scale in a curvelet domain and the supremum norm.
For a convenient calculation, we introduce a variable $\Be$ as a residual $\Beps$ in a known domain $\Omega \backslash D$.
The explanation for using the supremum norm of $\Beps$ in the curvelet domain can be found in (\ref{eq:sub:e:solution}). 
In particular, there are two terms in (\ref{eq:sub:e:solution}), including
a residual in $\Omega \backslash D$ with an updated Lagrange multiplier $\boldsymbol{\lambda_5}$,
i.e. $\Big( \chi_D^c \cdot^\times \Beps - \frac{\boldsymbol{\lambda_5}}{\beta_5} \Big)$,
and its curvelet smoothing term at a level $\nu$, i.e. $\CST \Big( \chi_D^c \cdot^\times \Beps - \frac{\boldsymbol{\lambda_5}}{\beta_5} \,, \nu \Big)$.
Assume at an iteration $t$, there are some remaining signals, e.g. texture, in residual 
$\Beps$ in $\Omega \backslash D$.
The curvelet soft-thresholding operator $\CST(\cdot, \cdot)$ reduces noise (or small scale objects) 
in $\big( \chi_D^c \cdot^\times \Beps - \frac{\boldsymbol{\lambda_5}}{\beta_5} \big)$ 
at a level $\nu$ in different scales and orientations.
By a subtraction operator in (\ref{eq:sub:e:solution}), $\Be$ at a iteration $t$ contains mainly noise or
small scale objects, see Figure \ref{fig:L_infty_Residual}.
\setcounter{subfigure}{0}

\subsection{The ``$\Beps$-problem'':} 
\begin{figure}[ht] 
\begin{center}
 
 \vspace*{-0.2\textwidth}
 \hspace*{-0.3\textwidth}
 \subfigure{ \includegraphics[width=1.6\textwidth]{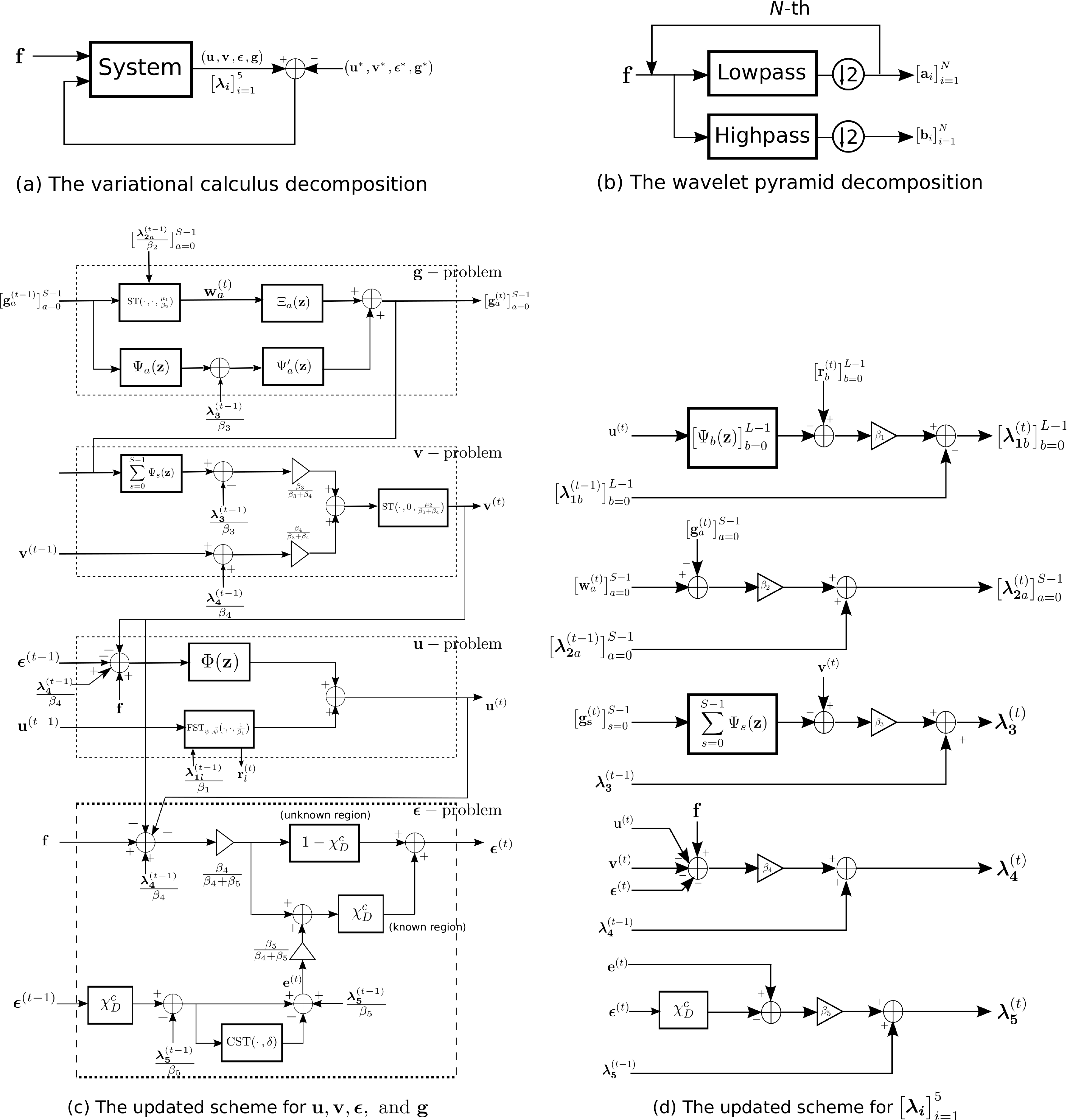} } 
 
 \caption{The pyramid scheme for the DG3PD inpainting model.}
 \label{fig:VariationPyramid}
\end{center}
\end{figure}

\begin{figure}[ht] 
\begin{center}
 \vspace*{-0.05\textwidth}
 
 \includegraphics[width=1\textwidth]{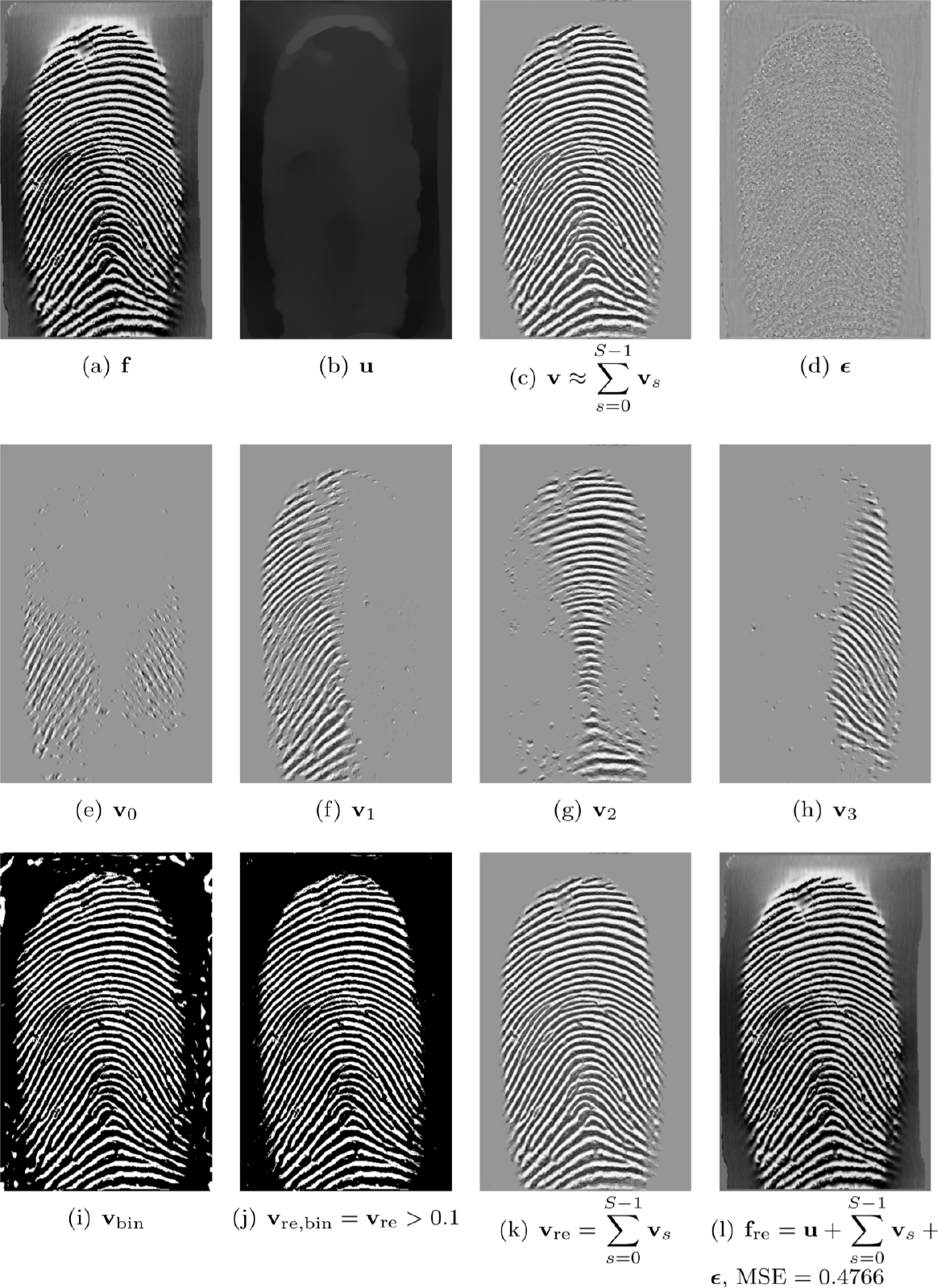}
 
 \caption{To test the directional property of texture and the piece-wise smooth cartoon in a decomposition,
          we apply DG3PD \cite{ThaiGottschlich2015DG3PD} for a fingerprint image. 
          Parameters: $\beta_4 = 0.03 \,, \theta = 0.9, L = 100 \,, S = 4 \,, c_1 = 1 \,, c_2 = 1.3 \,, c_{\mu_1} = c_{\mu_2} = 0.03 \,, \text{Iter} = 200$ 
          and $\nu = 15$.
          Cartoon $\Bu$ is similar to lowpass signal and directional texture $\big[\Bv_s\big]_{s=0}^3$ is directional bandpass signals. 
          The traditional linear filtering is difficult to design a filter to obtain a smooth lowpass signal with sharp edge as (b),
          because it is difficult to define frequencies (i.e. location in the Fourier domain) and their magnitude
          to approximate a cartoon-like image, see Figure \ref{fig:DG3PD:Fingerprint:Filter}.}
 \label{fig:DG3PD:Fingerprint}
\end{center}
\end{figure}

\begin{figure}[ht] 
\begin{center}
 \hspace*{-0.25\textwidth}
 \includegraphics[width=1.5\textwidth]{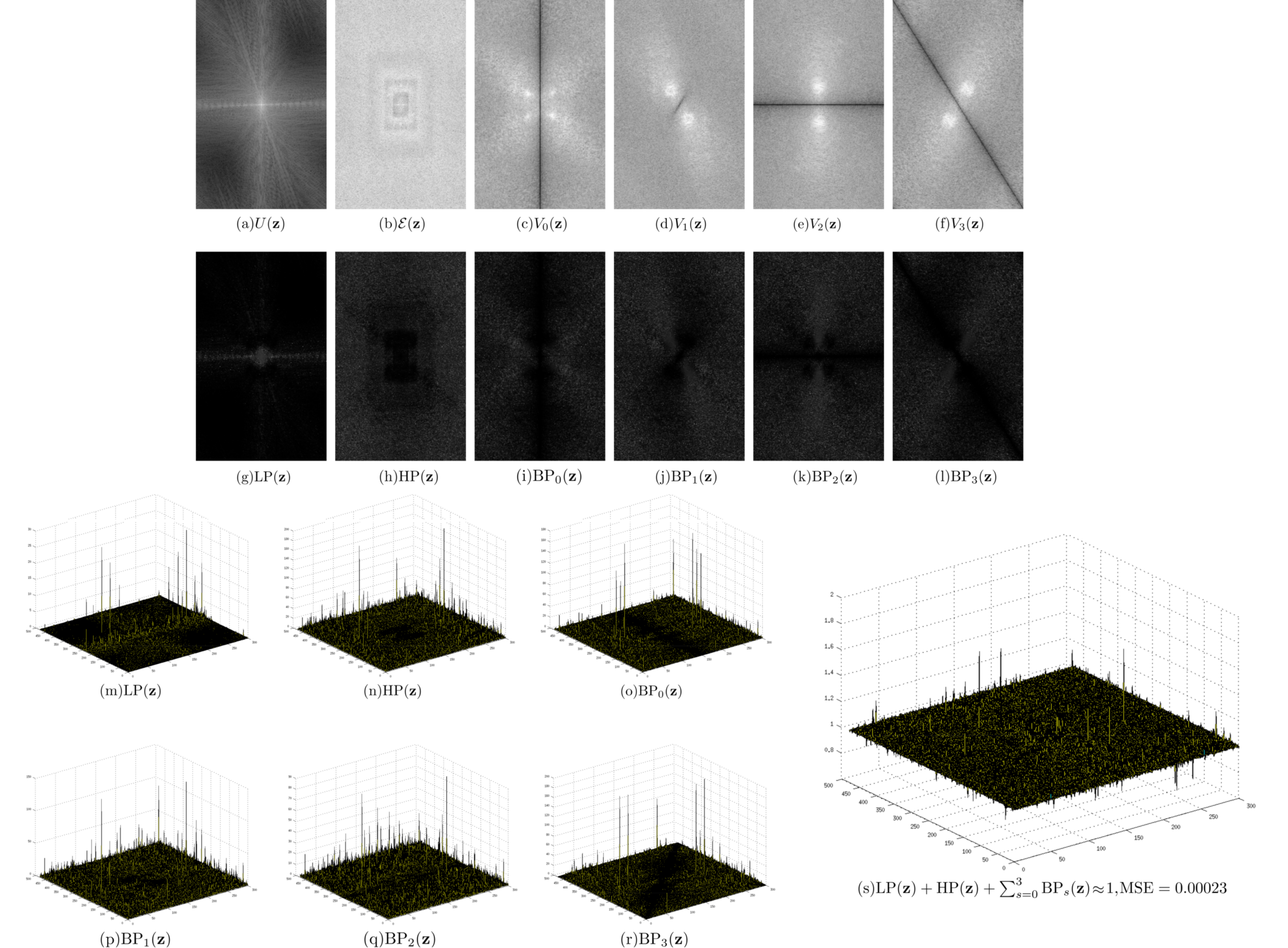}
  
 \caption{The first row shows the Fourier spectra of $\Bu, \Beps$ and $\big[\Bv_i\big]_{i=0}^3$ in Figure \ref{fig:DG3PD:Fingerprint}
          obtained by DG3PD \cite{ThaiGottschlich2015DG3PD}.
          Given a spectrum $F(\Bz)$ of the original image, their corresponding lowpass, highpas and directional
          bandpass in the second row are defined as $\text{LP}(\Bz) = \frac{U(\Bz)}{F(\Bz)} \,, \text{HP}(\Bz) = \frac{\mathcal E(\Bz)}{F(\Bz)}$
          and $\big[ \text{BP}_s(\Bz) \big]_{s=0}^3 = \frac{ \big[ V_{\text{re}s}(\Bz) \big]_{s=0}^3 }{ F(\Bz) }$,
          see (m)-(r) for their 3-dimensional filters.
          The error in (s) is due to the iterative method in ALM. In theory, when the number of iteration
          goes to infinity, a solution of a decomposition satisfies the condition of the perfect reconstruction.
          Given $N = 200$ iterations, MSE for the unity condition of all filters is small enough, i.e. it closely satisfies the unity property
          and the reconstructed signal well approximates to the original one.
         }
 \label{fig:DG3PD:Fingerprint:Filter}
\end{center}
\end{figure}

\begin{figure}[ht] 
\begin{center}
 \includegraphics[width=0.8\textwidth]{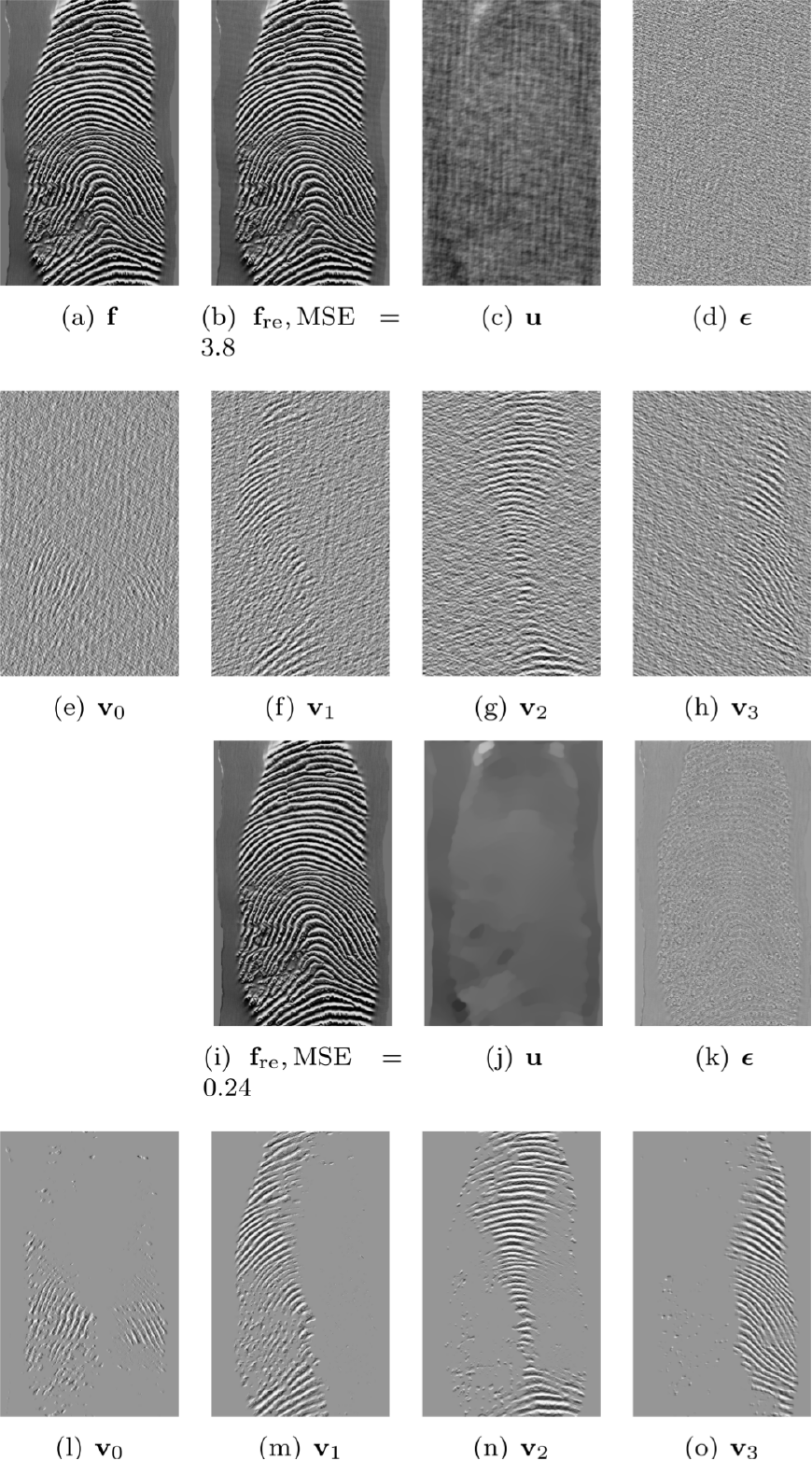}

 \caption{The first and second rows are the decomposition by DG3PD applied the same filters 
          as in Figure \ref{fig:DG3PD:Fingerprint:Filter} (in a linear convolution)
          for a similar fingerprint image as in Figure \ref{fig:DG3PD:Fingerprint}.
          We observe that applying the same linear filter for a ``similar'' image cannot give a good result for the decomposition.
          Thus, the linear filters are ``unique'' for every image to obtain a suitable decomposition.
          We apply DG3PD with an iterative method to obtain the decomposition in the third and fourth rows with the same parameters as in 
          Figure \ref{fig:DG3PD:Fingerprint:Filter}
          and MSE for the reconstructed image (i) is much smaller than (b), 
          see Figure \ref{fig:Fourier:uniqueFilter:3D} for its 3-dimensional linear filters.
          }
 \label{fig:Fourier:uniqueFilter}
\end{center}
\end{figure}

\begin{figure}[ht] 
\begin{center}
 \includegraphics[width=1\textwidth]{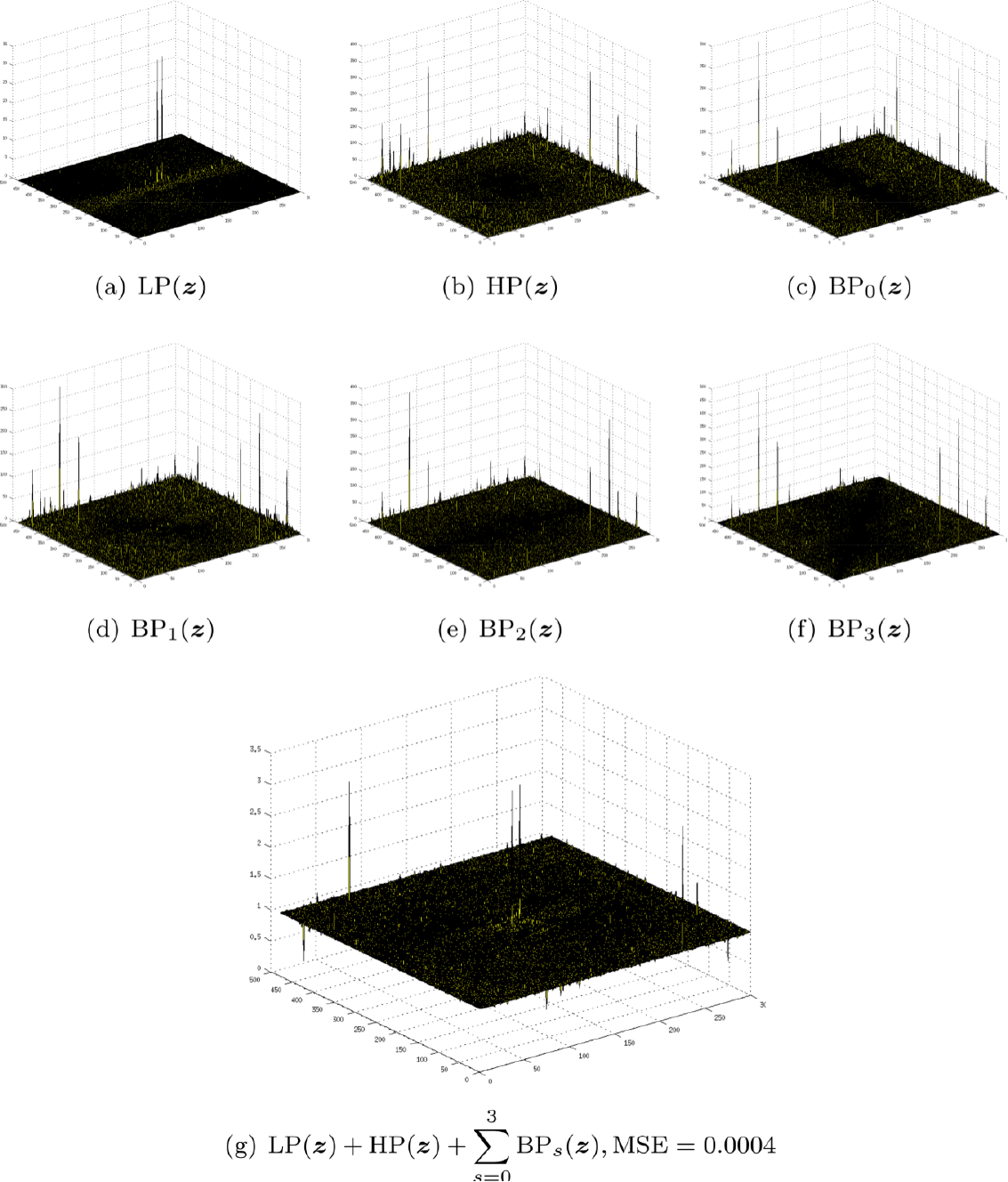}
 
 \caption{This figure shows the 3-dimensional filters for the fingerprint image in Figure \ref{fig:Fourier:uniqueFilter} (a)
          by DG3PD with the same parameters as in Figure \ref{fig:DG3PD:Fingerprint}.
          The linear filters are different and unique (in term of computation) to obtain a ``good'' approximated solution.   
          The scheme of filtering design depends on the characteristic of the image. 
          The variational method will automatically design a filter in a ``nonlinear way'' 
          (due to a minimization via an iterative method) and adapt to the characteristic of each image.
         }
 \label{fig:Fourier:uniqueFilter:3D}
\end{center}
\end{figure}

\begin{figure}[ht] 
\begin{center}
 \includegraphics[width=1\textwidth]{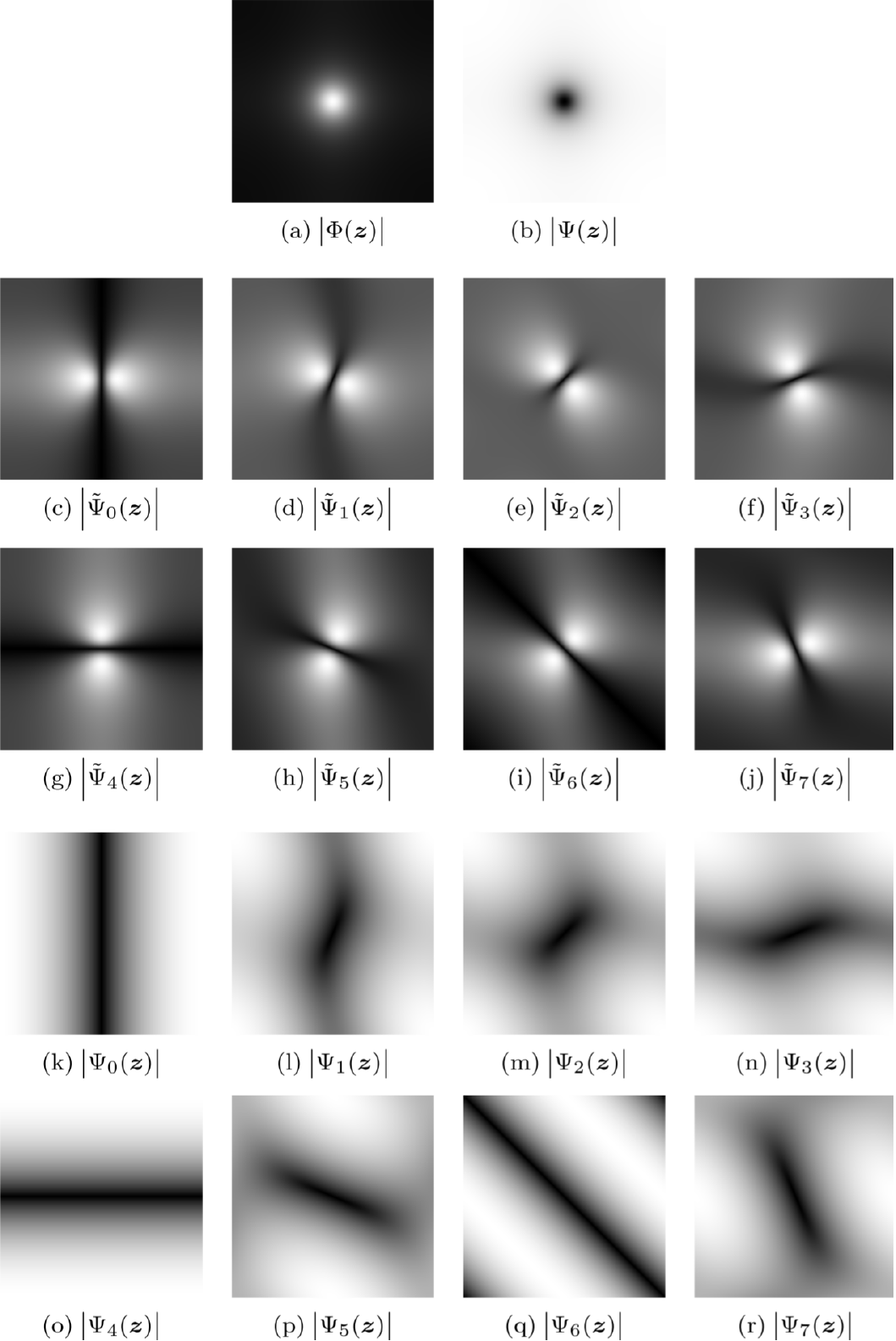}
  
 \caption{This figure shows the spectrum of the frame functions in the $\Bu$ and $\Bg$ problems with the parameters
          $L = S = 8 \,, \beta_1 = \beta_4 = 0.04 \,, \beta_2 = \beta_3 = 0.3$ and 
          $\displaystyle \Psi(\Bz) = \sum_{l=0}^{L-1} \tilde \Psi_l(\Bz) \Psi_l(\Bz)$,
          see Figure \ref{fig:FourierGa_L8} for more directional frame functions of the $\Bg$ problem.}
          
 \label{fig:FourierU_L8}
\end{center}
\end{figure}

\begin{figure}[ht] 
\begin{center}
 \includegraphics[width=1\textwidth]{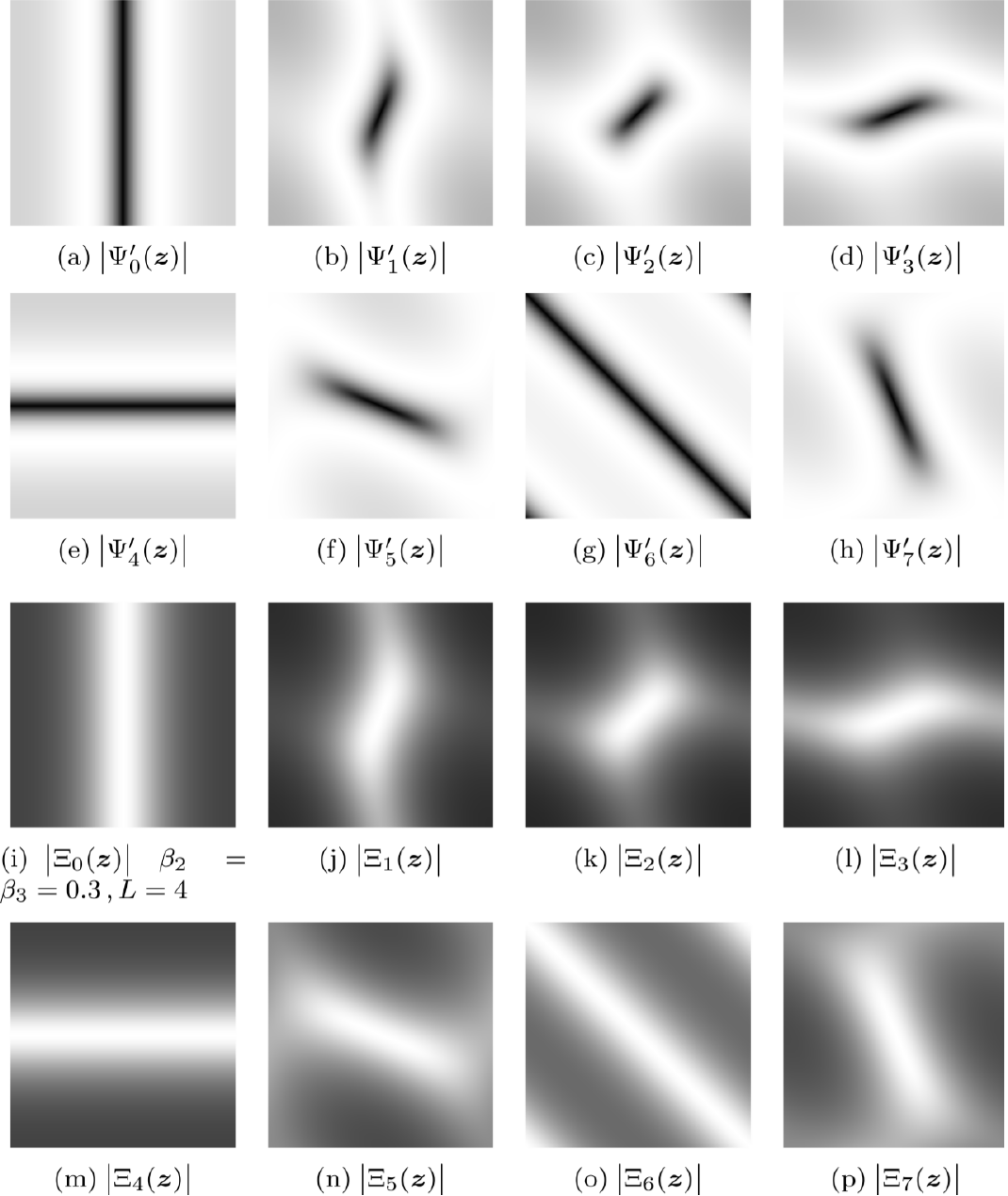}
   
 \caption{The spectrum of the frame functions in the $\Bg$-problem with $S=8$ directions.}
 \label{fig:FourierGa_L8}
\end{center}
\end{figure}

At iteration $t$, we denote the updated residual by the unity condition as 
$$
\Beps^{(t)}_\text{unity} = \Bf - \Bu^{(t)} - \Bv^{(t)} + \frac{\boldsymbol{\lambda_4}^{(t-1)}}{\beta_4} \,,
$$
and rewrite (\ref{eq:sub:epsilon:solution}) with the indicator functions 
on unknown domain $D$ and known domain $\Omega \backslash D$ as
\begin{equation*} 
 \Beps^{(t)} = \underbrace{ \Big[ \frac{\beta_4}{\beta_4 + \beta_5} \Beps^{(t)}_\text{unity}
 + \frac{\beta_5}{\beta_4 + \beta_5} \big( \Be^{(t)} + \frac{\boldsymbol{\lambda}_{\boldsymbol 5}^{(t-1)}}{\beta_5} \big) 
 \Big] }_{ :=~ \Beps^{(t)}_\text{known}}
 \cdot^\times \chi_D^c
 ~+~ \underbrace{ \frac{\beta_4}{\beta_4 + \beta_5} \Beps^{(t)}_\text{unity} }
     _{:=~ \Beps^{(t)}_\text{unknown}} \cdot^\times (1 - \chi_D^c).
\end{equation*}
We see that there are two terms in the residual $\Beps^{(t)}$, including the updated term $\Beps^{(t)}_\text{known}$
for $\Omega \backslash D$ and $\Beps^{(t)}_\text{unknown}$ for $D$.
And there are other two terms for updating $\Beps^{(t)}_\text{known}$ in $\Omega \backslash D$,
namely $\Beps^{(t)}_\text{unity}$ (from an unity condition) and 
$\Be^{(t)}$ (from a curvelet smoothing operator).
In a contrast to an open loop in a pyramidal decomposition (e.g. Laplacian pyramid, wavelet, curvelet, etc),
the solution of DG3PD inpainting can be considered as a closed loop of the pyramid scheme 
with the true solution $(\Bu^* \,, \Bv^* \,, \Beps^* \,, \Bg^* )$ in the minimization problem.
The loop of the updated Lagrange multipliers can be seen as a refinement. 
If we remove this loop from the scheme in Figure \ref{fig:VariationPyramid} (a),
the minimization in DG3PD becomes the quadratic penalty method. 
The removal would cause an imperfect reconstruction for the constraints, e.g. $\Bu + \Bv + \Beps \neq \Bf$,
see Figure 7 (p) and (q) in~\cite{ThaiGottschlich2015DG3PD}.
We note that the lowpass, bandpass, and highpass filters
for cartoon, texture and residual components are 'custom-made' for each image, 
resulting in different filters for different images, 
see Figure \ref{fig:Fourier:uniqueFilter:3D} 
and \ref{fig:Fourier:uniqueFilter}.
The directional properties of texture and their directional filter banks are illustrated 
in Figure \ref{fig:DG3PD:Fingerprint} 
and \ref{fig:DG3PD:Fingerprint:Filter}.

\setcounter{subfigure}{0}


\setcounter{subfigure}{0}

\setcounter{subfigure}{0}

\setcounter{subfigure}{0}

\setcounter{subfigure}{0}

\setcounter{subfigure}{0}

\setcounter{subfigure}{0}

\section{Conclusion} \label{sec:conclusion}

In this paper, we have addressed the very challenging task 
of restoring images with a large number of missing pixels, 
whose existing pixels are corrupted by noise and importantly,
the image to be restored contains both cartoon and texture elements.
The proposed DG3PD model for inpainting and denoising 
can cope with this threefold difficult problem. 
The task of simultaneous inpainting and denoising for cartoon and texture components
is solved by DG3PD decomposition, followed by inpainting and denoising both components separately
and finally, image restoration by synthesis of the restored components.
More specifically, the DG3PD inpainting model is based on a regularization in the Banach space in a 
discrete setting which is solved by ALM and ADMM.

In summary, the decomposition step of the proposed method has two major advantages 
for tackling this challenging problem.
Firstly, the decomposition step denoises simultaneously the cartoon and texture component.
Secondly, it allows to handle cartoon and texture in different ways.
The cartoon image is inpainted by DG3PD as described in Section \ref{sec:InpaintingDG3PD} and \ref{sec:solutionDG3PD}.
Separately, the texture component is inpainted followed by further denoising as described in Section~\ref{sec:textureInpainting}.
Therefore, the proposed DG3PD decomposition can also be understood as 
breaking the full problem down into 'smaller' subproblems (e.g. texture only inpainting) which are easier to solve.

Image restoration (or image denoising and inpainting) 
can be understood as an inverse problem
and it can described by a discrete innovation model (see Figure \ref{fig:DG3PD_inpainting:DiscreteInnovationModel}).
The assumption in this procedure is that signal has a sparse representation in suitable transform domains.
It is known from the probability theory that the sparsity (by $\ell_1$) is connected to
the Laplace distribution of the signal which results in the heavy tail distribution.
Note that the Laplace distribution with the $\ell_1$ norm is one of the best approximation
of the sparsity by $\ell_0$ norm.

Moreover, by choosing the priors and posterior according to the Bayesian framework and a maximum of a posterior (MAP),
we can understand the selection of parameters (see Figure \ref{fig:DG3PD:densityFunc} 
for the results of the selection of the heavy-tailed distribution, e.g. the Laplace distribution).
Note that MAP requires an assumption on the noise, e.g. Gaussian or Laplacian, to establish a minimization.
In our proposed model, there is no requirement for an assumption on noise distribution, 
e.g. independent identical distributed (i.i.d), or correlated and Gaussian or non-Gaussian (due to the measurement $\norm{ \cC \{ \cdot \} }_{\ell_\infty}$,
which is similar to the Dantzig selector \cite{CandesTao2007}).

\section*{Acknowledgements}

D.H.~Thai is supported by the National Science Foundation 
under Grant DMS-1127914 to the Statistical and Applied Mathematical Sciences Institute.\\
C.~Gottschlich gratefully acknowledges the support of the 
Felix-Bernstein-Institute for Mathematical Statistics in the Biosciences 
and the Niedersachsen Vorab of the Volkswagen Foundation.

\end{document}